\documentclass[preprint,11pt]{elsarticle}  
\usepackage[a4paper,margin=2.5cm]{geometry}  

\usepackage[utf8]{inputenc}
\usepackage[T1]{fontenc}
\usepackage{graphicx}
\usepackage{booktabs}
\usepackage{multirow}
\usepackage{amsmath}
\usepackage{amssymb}
\usepackage{xcolor}
\usepackage{adjustbox}
\usepackage{microtype}
\usepackage[hidelinks]{hyperref}
\usepackage{xurl}  

\usepackage[section]{placeins}
\newcommand{\pairwidth}{\textwidth}\newcommand{\pairsep}{\par\vspace{1.5em}}


\newcommand{\CapTabMax}{0.78\textheight}

\journal{Artificial Intelligence in Medicine}

\makeatletter
\def\ps@pprintTitle{%
  \let\@oddhead\@empty\let\@evenhead\@empty
  \let\@oddfoot\@empty\let\@evenfoot\@empty}
\makeatother

\newcommand{\paperabstract}{%
Foundation models pretrained on large natural-image collections are increasingly adapted for medical image segmentation, in the expectation that they will replace models trained for each task. Most comparisons, however, use data from the training distribution and a single task-specific baseline whose configuration is rarely justified, and therefore provide limited evidence of performance at new clinical sites. It also remains unclear how much of a foundation model must be adapted, and what determines this. We adapted a foundation model, DINOv2, with strategies ranging from tuning only its bias terms to full fine-tuning, and compared it with U-Nets and ResU-Nets trained from scratch. Experiments covered colonoscopy, breast ultrasound and dermoscopy, with testing on held-out data from the training source and on independent datasets from other institutions. The adapted foundation model generalised better on every external dataset. Within the training distribution, the task-specific models remained competitive when training data were abundant, and delineated boundaries more precisely, while the foundation model more often missed small lesions. Only a small fraction of the foundation model required adaptation, and the capacity needed was determined by the imaging domain rather than by the adaptation method. A widely used polyp test set was also found to duplicate images from another benchmark. Foundation-model adaptation is therefore warranted where models must work across sites, with capacity selected on the target domain.
}
\newcommand{\paperkeywords}{medical image segmentation \sep foundation models \sep parameter-efficient fine-tuning \sep cross-dataset generalisation \sep U-Net \sep ResU-Net \sep nnU-Net}

\begin{document}

\begin{frontmatter}

\title{Medical Image Segmentation: Foundation Model Adaptation versus Task-Specific Training}

\author[brunel]{Emre Kutlusan}
\author[brunel]{Yongmin Li\corref{cor1}}
\ead{yongmin.li@brunel.ac.uk}
\cortext[cor1]{Corresponding author.}

\affiliation[brunel]{organization={Department of Computer Science, Brunel University of London},
            city={Uxbridge},
            postcode={UB8 3PH},
            country={United Kingdom}}

\begin{abstract}
\paperabstract
\end{abstract}


\end{frontmatter}

\section{Introduction}\label{sec:intro}

Vision foundation models, pretrained on hundreds of millions of images at a scale no clinical dataset approaches, have made a question urgent for medical image segmentation. Given a new dataset and task, is it better to adapt one of these models, or to train a task-specific architecture from random initialisation as the field has done since U-Net?

The question matters because of where segmentation models are deployed. A model trained at one hospital is intended for use at others, on images from different scanners, protocols and patient populations. This distribution shift, rather than in-domain accuracy, is the principal obstacle to clinical adoption, and it is invisible in the measurement usually reported. Every model in this study is therefore evaluated under two regimes: in domain, on held-out data from the training distribution, and out of distribution, on data from institutions that contributed nothing to training.

The literature comparing the two approaches generally reports that adapted foundation models win, and three properties of how that comparison is made limit what follows from it. It is usually conducted within the training distribution, so a property of the evaluation is reported as a property of the models. It is usually made against a single task-specific model whose configuration is not itself defended, and \citet{isensee2024revisited} show that many claimed improvements over U-Net are made against a weakly configured instance of it; the two objections compound, since an advantage so obtained can be an artefact of both at once. It is also usually made at a single operating point, that is, at one choice of settings such as the adaptation rank or the training budget, so a statement about one configuration appears to be a statement about the methods.

A parallel gap concerns how much adaptation a target requires. \citet{dutt2024peft}, in classification and generation tasks, report that parameter-efficient methods are most effective in low-data regimes, making training-set size the governing quantity and one known before any experiment; \citet{teuber2026peft} report the contrary directly, finding no such advantage even at very small training sizes, and attribute the variation to properties of the dataset, which are not known in advance. What neither establishes is whether the requirement belongs to the domain or to the method supplying the adaptation.

The aim of this study is to determine, for medical image segmentation, when adaptation of a pretrained foundation model is warranted in place of a task-specific architecture, and what governs the amount of adaptation such a model requires. It does so along two axes. The architecture axis compares an adapted foundation model with task-specific architectures trained from random initialisation; the adaptation axis varies how much of the foundation model is adapted, measuring how the choice of adaptation strategy affects cross-dataset generalisation and whether the capacity required is a property of the domain or of the method supplying it.

Both axes span three imaging domains---colonoscopy polyp imaging, breast ultrasound and skin lesion dermoscopy---each with one training source and held-out test sets, giving eight out-of-distribution test sets, one of which (CVC-T) consists of images also present in another (CVC-ColonDB; Section~\ref{sec:datasets}). The foundation model is a DINOv2-Small encoder with a minimal linear decoder, adapted under strategies spanning 385 to 22.1 million trainable parameters. The task-specific comparators are a plain U-Net and a residual U-Net, both under the nnU-Net recipe and each tuned independently.

The contributions are threefold:
\begin{enumerate}
\item A controlled comparison of foundation-model adaptation against task-specific training from scratch, with the evaluation regimes separated throughout and against baselines shown to be competitive with published architectures under an identical protocol: adaptation leads on all eight out-of-distribution test sets in DSC, IoU, sensitivity and HD95 under both architectures and its advantage grows as training data shrinks within a domain, most clearly in polyp, while in domain at the full training set no accuracy advantage was observed.
\item Empirical findings on adaptation capacity: the step from a frozen encoder to bias-only tuning captures most of the available benefit in polyp and breast, LoRA at 2.7 per cent of the parameters is equal or ahead of full fine-tuning in DSC on all five polyp sets across three seeds, and the capacity a domain requires is a property of the domain rather than of the method supplying it.
\item Practical recommendations for choosing between and evaluating the two approaches: when adaptation is warranted, how much capacity to use and how to select it, which failure modes to check before deployment, and which evaluation hazards in standard benchmarks to control.
\end{enumerate}

\section{Related work}\label{sec:related}

Medical image segmentation spans a wide range of imaging modalities, including X-ray radiography \citep{candemir2014lung}, computed tomography (CT) \citep{heller2021kits19,antonelli2022msd}, magnetic resonance imaging (MRI) \citep{menze2015brats,payette2025feta}, ultrasound \citep{aldhabyani2020busi}, colonoscopy \citep{bernal2015wm,jha2020kvasir}, dermoscopy \citep{codella2018isic,eltayef2017pigment,eltayef2017melanoma}, fundus photography \citep{salazar2011opticdisc,kaba2014vessels} and optical coherence tomography (OCT) \citep{dodo2019annotation,ndipenoch2024oct}. These modalities differ in how the image is formed, in the appearance and scale of the anatomical targets and in how much annotated data is available, and each has accumulated its own methods and benchmarks. The three domains studied here, colonoscopy, breast ultrasound and dermoscopy, are drawn from this range, and the remainder of this section reviews the segmentation architectures, foundation models, adaptation methods and evaluation practices on which the study builds.

\subsection{Segmentation architectures and nnU-Net}\label{sec:rw-nnunet}

Before deep learning, segmentation methods were typically designed around the anatomy and appearance of a single task, for example graph cuts for retinal vessels \citep{salazar2010vessel} and for retinal layers in OCT \citep{dodo2018graphcut,dodo2019layers}, and Gaussian mixture and other probabilistic models for vessel extraction \citep{kaba2013gmm,kaba2014vessels}. Learned encoder--decoder networks have since largely replaced such pipelines, although task-specific deep architectures continue to be proposed for individual targets such as colonoscopy polyps \citep{fan2020pranet,zhang2020acsnet,huang2021hardnetmseg,dong2023polyppvt}. The U-Net of \citet{ronneberger2015unet} still dominates medical segmentation, and of the many variants proposed since, \citet{mcconnell2023nnunet}, comparing residual, dense, inception and attention families across eight datasets, conclude that none dominates and that the optimal choice is dataset-dependent. nnU-Net \citep{isensee2021nnunet} reframed how such comparisons should be conducted by proposing a procedure rather than an architecture: it derives the network topology, patch size, resampling strategy, batch size and training schedule from a fingerprint of the dataset, and its central empirical claim, that a U-Net so configured matches or exceeds most published architectural modifications, makes the configuration of a baseline a methodological question rather than an implementation detail. That claim was sharpened in nnU-Net Revisited \citep{isensee2024revisited}, which finds that many papers claiming to outperform U-Net do so against a weakly configured instance of it, and which also reports strong performance from residual-encoder nnU-Net configurations, evaluated on three-dimensional multi-class segmentation across large datasets. Both results shape the design here: a residual variant, in the spirit of those configurations but with residual blocks in both encoder and decoder, was adopted as the primary task-specific baseline, and the 2015 architecture was deliberately not used, since employing it would have reproduced the weak-baseline error in the direction that flatters the conclusions drawn here.

\subsection{Foundation models and their medical adaptation}\label{sec:rw-fm}

DINOv2 \citep{oquab2024dinov2} is trained self-supervised on 142 million curated natural images with no labels, and its patch-level features transfer to dense prediction without task-specific supervision during pretraining. Because it has never seen a segmentation mask, whether its representations transfer to medical segmentation is a question about representational generality rather than about label overlap, which is why it was selected here. Its medical transfer has been assessed by \citet{baharoon2023dinov2} across nine radiology benchmarks and by \citet{ayzenberg2024dinov2} in few-shot segmentation, and \citet{li2025dinounet} report that pairing DINO features with a U-Net decoder supplies a strong inductive prior for medical segmentation.

The second family originates with the Segment Anything Model \citep{kirillov2023sam}, trained on a billion masks over eleven million images, and is adapted to medical images by retraining on large-scale medical data \citep{ma2024medsam}, by applying LoRA to the image encoder \citep{zhang2023samed}, by large-scale two-dimensional adaptation \citep{cheng2023sammed2d} or by inserting adapters into a frozen backbone \citep{wu2025medsamadapter}; SAM-Mamba \citep{dutta2025sammamba} reports strong cross-dataset polyp performance and appears in the comparison of Section~\ref{sec:arch}. A model supervised on segmentation masks and a model that has never seen one require different amounts of adaptation to reach a medical target, so a finding about how much adaptation one needs does not automatically transfer to the other (Section~\ref{sec:relation}).

\subsection{Parameter-efficient adaptation}\label{sec:rw-peft}

Parameter-efficient methods update a small fraction of a pretrained model's parameters, avoiding the memory and storage cost of full fine-tuning and, on small target datasets, the risk of destabilising representations that took large-scale pretraining to acquire. Adapter modules \citep{houlsby2019adapters} and visual prompt tuning \citep{jia2022vpt} add computation at inference, which is the property LoRA avoids. LoRA \citep{hu2022lora} freezes the pretrained weights and adds a trainable low-rank decomposition to selected projections, so that the effective weight update is constrained to rank $r$; the rank controls the capacity of the update and is the method's principal hyperparameter, and because the adapters can be merged into the base weights after training, inference cost is unchanged. \citet{kalajdzievski2023rslora} shows that the conventional scaling factor slows learning and stunts performance at higher ranks and proposes a rank-stabilised alternative, which is why the present work holds the scaling constant so that the rank sweep of Section~\ref{sec:capacity} varies rank and nothing else. BitFit \citep{benzaken2022bitfit} freezes all weight matrices and trains only bias terms, under one tenth of one per cent of the parameters in the models it was proposed for, and its authors report performance competitive with full fine-tuning on language tasks with the advantage concentrated in low-data regimes. The absence of a rank parameter and its different intervention site make it the natural control for testing whether a capacity requirement is a property of low-rank factorisation or of the target, which is the role it serves in Section~\ref{sec:capacity}.

The literature applying these methods to medical imaging contains an unresolved disagreement, and it is the point at which this study enters. \citet{dutt2024peft} benchmark parameter-efficient methods across medical image analysis tasks and report an ordering that depends on data volume: in their HAM10000 classification experiment with ViT-Large, LoRA overtakes full fine-tuning from approximately eighty per cent of the available data downward, while BitFit is superior only below twenty per cent. \citet{teuber2026peft} evaluate parameter-efficient adaptation of SAM across biomedical datasets and report instead that the relative effectiveness of these methods differs by dataset in a way that data volume alone does not account for, and that for computational efficiency, where in the encoder a method intervenes matters more than which method it is. The two are not straightforwardly reconcilable, and no study separates them, which would require varying data volume within a domain while comparing across domains at matched volume. Section~\ref{sec:relation} returns to the disagreement with results in hand.

\subsection{Cross-dataset evaluation and data hygiene}\label{sec:rw-xdata}

Most published results are obtained on held-out data from the training source, which is not the deployment setting, and under a consistent protocol well-tuned empirical risk minimisation is competitive with most proposed domain generalisation algorithms \citep{gulrajani2021domainbed}. The protocol of \citet{fan2020pranet} partitions five polyp collections into a fixed training portion and three held-out sets, and its widespread adoption means published scores are comparable in experimental setup rather than merely in metric, which is what permits the baselines here to be positioned against the literature. U-Bench \citep{tang2025ubench} evaluates 100 U-Net variants across 28 datasets and 10 modalities, reserving eight unseen datasets for zero-shot evaluation, and reports that in-domain metrics have largely saturated while zero-shot generalisation continues to separate methods. \citet{ndipenoch2024oct} apply the same distinction within medical segmentation, comparing an adapted SAM against nnU-Net variants across three OCT device vendors and reporting the task-specific models slightly ahead. A domain qualifies for such a study only if it offers a training source with independently acquired, comparably annotated, public test sets; three satisfy this and were chosen to differ in modality, by a factor of more than four in training-set size, and in severity of shift. Selecting domains that differ this way is what makes it possible to ask whether a regularity is a property of a method or of the data it was measured on, which a single-domain study cannot ask.

Cross-dataset evaluation is only as sound as the separation between the sets compared. \citet{truong2023duplicates} report that medical datasets often contain near-duplicate images that can bias evaluation, identifying possible near-duplicates in the widely used Medical Segmentation Decathlon and benchmarking their detection under transformations that include cropping and compression. On metrics, \citet{taha2015metrics} set out the consequences of choosing among them. On representations, centred kernel alignment \citep{kornblith2019cka} is the standard tool for comparing representations across networks, and \citet{nguyen2021wide} use the debiased estimator adopted in Section~\ref{sec:stability}.

\subsection{Research gaps}\label{sec:rw-gap}

Three gaps follow.
\begin{enumerate}
\item \textit{When adaptation should be preferred.} The literature does not settle when one approach should be preferred over the other, because comparisons are typically made within the training distribution and against a baseline whose configuration is not itself defended, the weakness nnU-Net \citep{isensee2021nnunet} and nnU-Net Revisited \citep{isensee2024revisited} identify in architectural comparisons.
\item \textit{How much adaptation a target requires.} Separating the data-volume effects and dataset-dependent outcomes of Section~\ref{sec:rw-peft} requires varying data volume within a domain while comparing across domains at matched volume, and testing the same question under two methods that share no mechanism, so that any regularity found cannot be a property of one method's parameterisation.
\item \textit{How far the answers depend on configuration.} nnU-Net Revisited \citep{isensee2024revisited} and U-Bench \citep{tang2025ubench} reach their conclusions by breadth, evaluating many methods under a common protocol, which leaves little room to vary the configuration of any one comparison.
\end{enumerate}
This study varies data volume on the architecture axis and compares domains and methods on the adaptation axis; the interaction between capacity requirement and data volume is left open (Section~\ref{sec:limitations}). For the third gap it takes the complementary trade-off, few methods examined at many operating points, which is the reason it covers three domains and two adaptation methods, examining their sensitivity to adaptation rank, training budget and evaluation regime.

\section{Methods}\label{sec:methods}

\subsection{Design and scope}\label{sec:design}

The foundation model adopted is DINOv2 \citep{oquab2024dinov2}, in its smallest variant. It was chosen deliberately in preference to segmentation-specific or medical foundation models such as SAM \citep{kirillov2023sam}, which is trained on a billion masks, and SAMed \citep{zhang2023samed}, which fine-tunes SAM on medical images. The aim is to demonstrate the effectiveness of the adaptation approach with a small, general-purpose model that has never been trained on segmentation masks or medical images, so that any advantage it shows can be attributed to adaptation rather than to prior exposure to the target task or domain. Such a model represents a conservative case: if it suffices, larger foundation models, and those pretrained on masks or medical data, would be expected to perform at least as well. A finding about how much adaptation one model family needs does not, however, automatically transfer to another (Section~\ref{sec:rw-fm}), and repeating the adaptation axis on SAM, which SAMed adapts with LoRA in the same way as the configurations here, is left to future work (Section~\ref{sec:limitations}).

\begin{figure*}[t]
\centering
\includegraphics[width=0.9\textwidth]{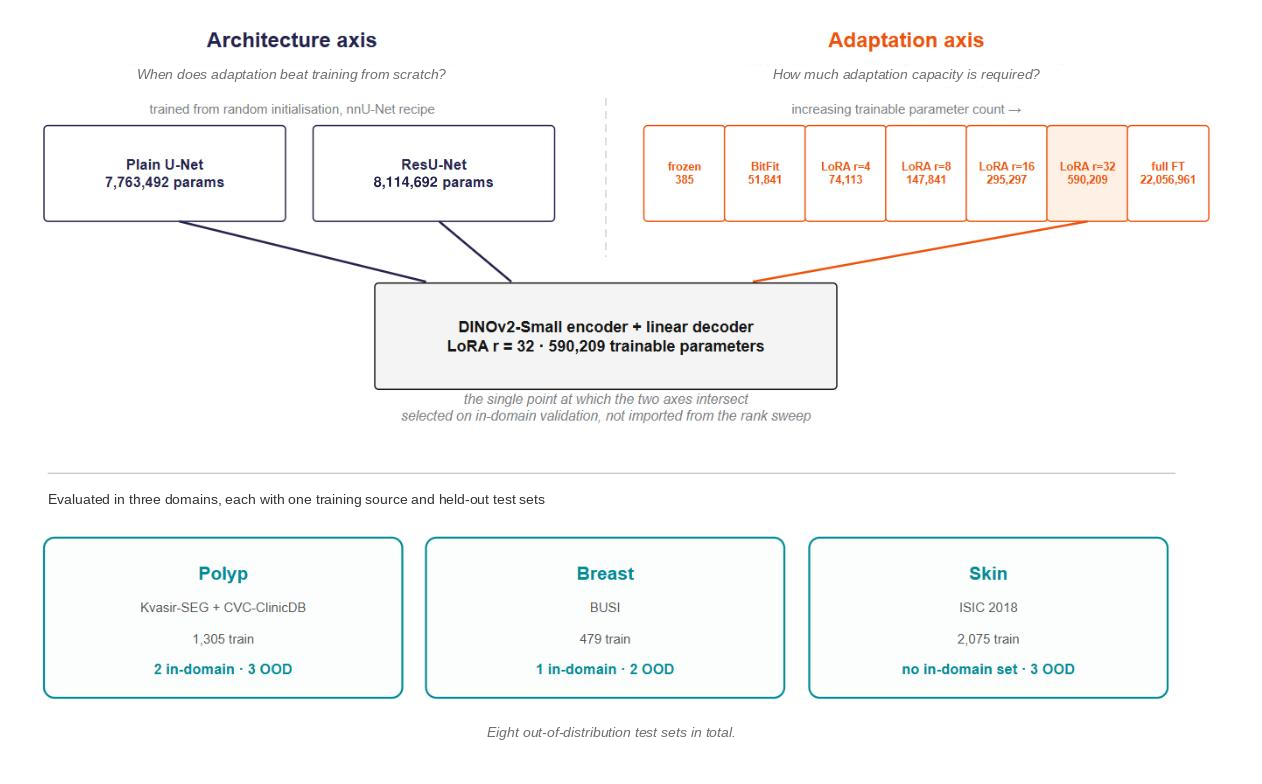}
\caption{Experimental design. The architecture axis compares an adapted DINOv2 against two models trained from random initialisation; the adaptation axis compares seven configurations of the same model by trainable parameter count. The two intersect at LoRA rank 32, selected on in-domain validation. All comparisons run across three domains and eight out-of-distribution test sets.}
\label{fig:design}
\end{figure*}

The design is organised around two axes kept separate throughout, summarised in Fig.~\ref{fig:design}. The architecture axis asks when adaptation outperforms task-specific training: an adapted DINOv2 encoder is compared against a plain U-Net and a residual U-Net (ResU-Net) trained from random initialisation. The adaptation axis asks how much adaptation is required: the foundation model is compared only against itself under strategies of increasing capacity. No task-specific model appears on the adaptation axis, and no adaptation strategy is applied to the task-specific architectures, which have no pretrained parameters to adapt.

The adaptation axis is conditional on the architecture axis, and the two intersect at a single point rather than as a factorial design. Only one DINOv2 configuration enters the architecture-axis comparison, selected on in-domain validation rather than imported from the rank sweep, and results from one axis are not used to support claims about the other. Claims on the adaptation axis consequently rest on a single backbone, a limitation returned to in Section~\ref{sec:limitations}.

All models share training splits, evaluation code and metric definitions. Augmentation and optimiser settings differ by model family, the task-specific models using a stronger pipeline than the adaptation experiments. Where budgets or learning rates differ it is because each configuration runs at its own selected setting rather than an arbitrary shared one, and each such difference is identified where it occurs. Where multi-seed results exist the seed-averaged value is used, which on the architecture axis affects the skin domain only.

The use of three domains is a methodological choice rather than a matter of scale. A recurring finding of this work is that a regularity observed in one domain does not hold in the next, which a single-domain study is structurally unable to detect.

\subsection{Domains, datasets and leakage control}\label{sec:datasets}

Three domains were selected to span imaging modality, training-set size and severity of distribution shift, for the reasons given in Section~\ref{sec:rw-xdata}. Each provides one training source and held-out test sets acquired by different institutions or under different protocols, with the exception reported below. After cleaning, the training and validation splits contain 1,305 and 145 images in polyp, 479 and 54 in breast, and 2,075 and 519 in skin (Table~\ref{tab:datasets}); test-set sizes are given in Table~\ref{tab:main}.

\begin{table*}[tbp]
\centering
\begin{minipage}[t]{\pairwidth}
\centering
\caption{Datasets by domain; counts are images after cleaning. Asterisks mark skin sets drawn from the same archive as the training data.}
\label{tab:datasets}
\footnotesize
\begin{tabular*}{\linewidth}{@{\extracolsep{\fill}}lp{0.33\linewidth}p{0.42\linewidth}@{}}
\toprule
Domain & Training source (train / val) & Test sets (images) \\
\midrule
Polyp  & Kvasir-SEG + CVC-ClinicDB (1,305 / 145) & In domain: Kvasir 100, CVC-ClinicDB 62. Out of distribution: CVC-ColonDB 380, ETIS 196, CVC-T 60 \\
\midrule
Breast & BUSI (479 / 54) & In domain: BUSI-val 54. Out of distribution: BrEaST 252, BUS-UCLM 260 \\
\midrule
Skin   & ISIC 2018 (2,075 / 519) & Out of distribution: ISIC 2016* 269, ISIC 2017* 172, PH2 200 \\
\bottomrule
\end{tabular*}
\end{minipage}\pairsep
\begin{minipage}[t]{\pairwidth}
\centering
\caption{Leakage detection by domain.}
\label{tab:leakage}
\footnotesize
\begin{tabular*}{\linewidth}{@{\extracolsep{\fill}}lp{0.22\linewidth}p{0.25\linewidth}p{0.28\linewidth}@{}}
\toprule
Domain & Method & Detects & Outcome \\
\midrule
Polyp  & md5 hash of image bytes & Byte-identical files & No overlap \\
\midrule
Skin   & Filename stem matching & Shared archive identifiers & Test sets filtered to 2016*, 2017* \\
\midrule
Breast & Perceptual hash & Near-duplicates, re-encodings & 114 removed (17.6\% of 647) \\
\bottomrule
\end{tabular*}
\end{minipage}
\end{table*}

Polyp combines Kvasir-SEG \citep{jha2020kvasir} and CVC-ClinicDB \citep{bernal2015wm} under the protocol of \citet{fan2020pranet}, with CVC-ColonDB \citep{bernal2012polyp}, ETIS-LaribPolypDB \citep{silva2014etis} and CVC-T \citep{vazquez2017benchmark} held out; the protocol was adopted because its widespread use makes published scores comparable in setup, which permits the task-specific baselines to be positioned against the literature in Section~\ref{sec:arch}. Breast uses BUSI \citep{aldhabyani2020busi} with BrEaST \citep{pawlowska2024breast} and BUS-UCLM \citep{vallez2025busuclm} held out, each acquired at a different institution, and four BUS-UCLM images with mismatched mask dimensions were excluded. Skin uses ISIC 2018 \citep{codella2019isic, tschandl2018ham} with filtered subsets of ISIC 2016 \citep{gutman2016isic} and ISIC 2017 \citep{codella2018isic} and the independent PH2 collection \citep{mendonca2013ph2}. Breast has no separate in-domain test set, so its validation split serves that role. In skin the validation split was used for model selection only and is not reported as an in-domain result, so skin contributes to the out-of-distribution comparison alone. One caveat is carried to the discussion: although held out entirely, ISIC 2016* and 2017* originate from the same archive as the training data and represent a milder shift than the polyp or breast sets, PH2 being the only genuinely independent skin source. All datasets used here are publicly available and were released for research use by their original providers, and no new human data were collected.

\begin{figure}[tbp]
\centering
\includegraphics[width=\linewidth]{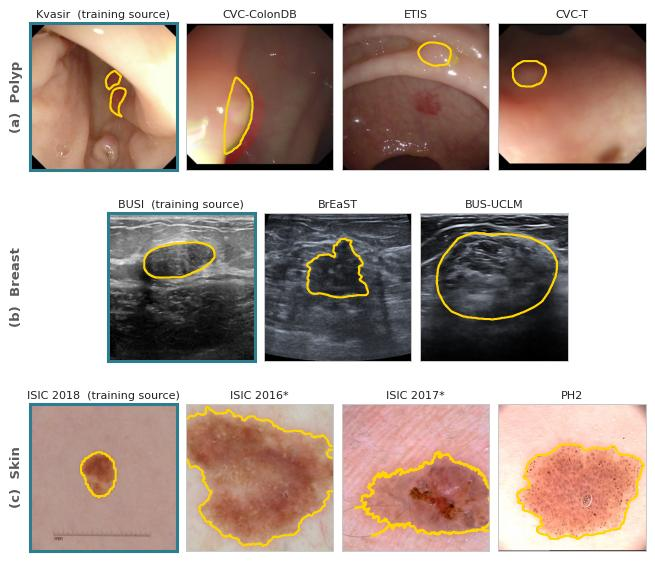}
\caption{Example cases from each domain. One image per dataset with its ground-truth contour in yellow; images are centre-cropped to a common square footprint and are otherwise unmodified. The training source of each domain is framed in colour, with one of polyp's two training sources shown, and every other set is held out. Breast has one fewer test set than the other two domains, since only two independent breast ultrasound sources with segmentation masks were available. The asterisk marks skin test sets drawn from the same archive as the training source.}
\label{fig:examples}
\end{figure}

Fig.~\ref{fig:examples} shows one annotated case from each training source alongside cases from that domain's out-of-distribution sets, illustrating the variation in illumination, field of view and annotation convention the cross-dataset evaluation measures.

Cross-dataset evaluation is meaningful only if the test sets are genuinely unseen. Where overlap goes undetected, out-of-distribution scores are inflated by an unquantifiable amount and the comparison this study rests on becomes uninterpretable, so each domain was examined before any experiment was run, with a method suited to its failure mode (Table~\ref{tab:leakage}).

In polyp the protocol partitions the sources into disjoint portions, so the residual risk is accidental duplication, and md5 hashing found no collisions; it is the weakest method, failing on any re-encoding, but adequate since the collections are distributed separately. In skin, because the ISIC archive preserves identifiers across releases, overlap was found by matching filename stems, and matches were removed from the test sets rather than the training set, preserving the training distribution. This renders the skin results non-comparable with published scores on the full ISIC test sets, accepted because an uncontaminated smaller set is preferable to a contaminated comparable one. Breast required the most preparation. Of 780 images, 133 are normal cases with empty masks and were removed, and the remaining 647 contain near-duplicates that md5 hashing largely misses: a perceptual hash at a Hamming distance of five or less placed 212 of them in 98 groups, only one of which contains byte-identical files, and keeping one image per group removed 114, leaving 533. Inspection of a sample of four groups showed the same acquisitions annotated twice with differing masks, so their presence would constitute a leak carrying contradictory labels. The count depends on the threshold, from 89 removed images at a distance of three to 137 at eight.

Checking the test sets against one another, which the protocol does not require, showed that two of them are not independent: all 60 CVC-T images are pixel-identical to images in CVC-ColonDB, as are 59 of their 60 masks, while none of the files is byte-identical, so the md5 check applied to the training data would not have detected the overlap. CVC-T is retained because it belongs to the standard protocol and to the published comparisons of Section~\ref{sec:arch}, but its results are not independent evidence alongside CVC-ColonDB, and polyp therefore contributes two independent out-of-distribution sources rather than three. The overlap is consistent with the provenance of the data: the EndoScene benchmark from which CVC-T is drawn defines its images as CVC-ColonDB (300 images from 13 sequences) together with CVC-ClinicDB (612 images) \citep{vazquez2017benchmark}.

\subsection{Adaptation strategies}\label{sec:adapt}

\begin{figure*}[t]
\centering
\includegraphics[width=\linewidth]{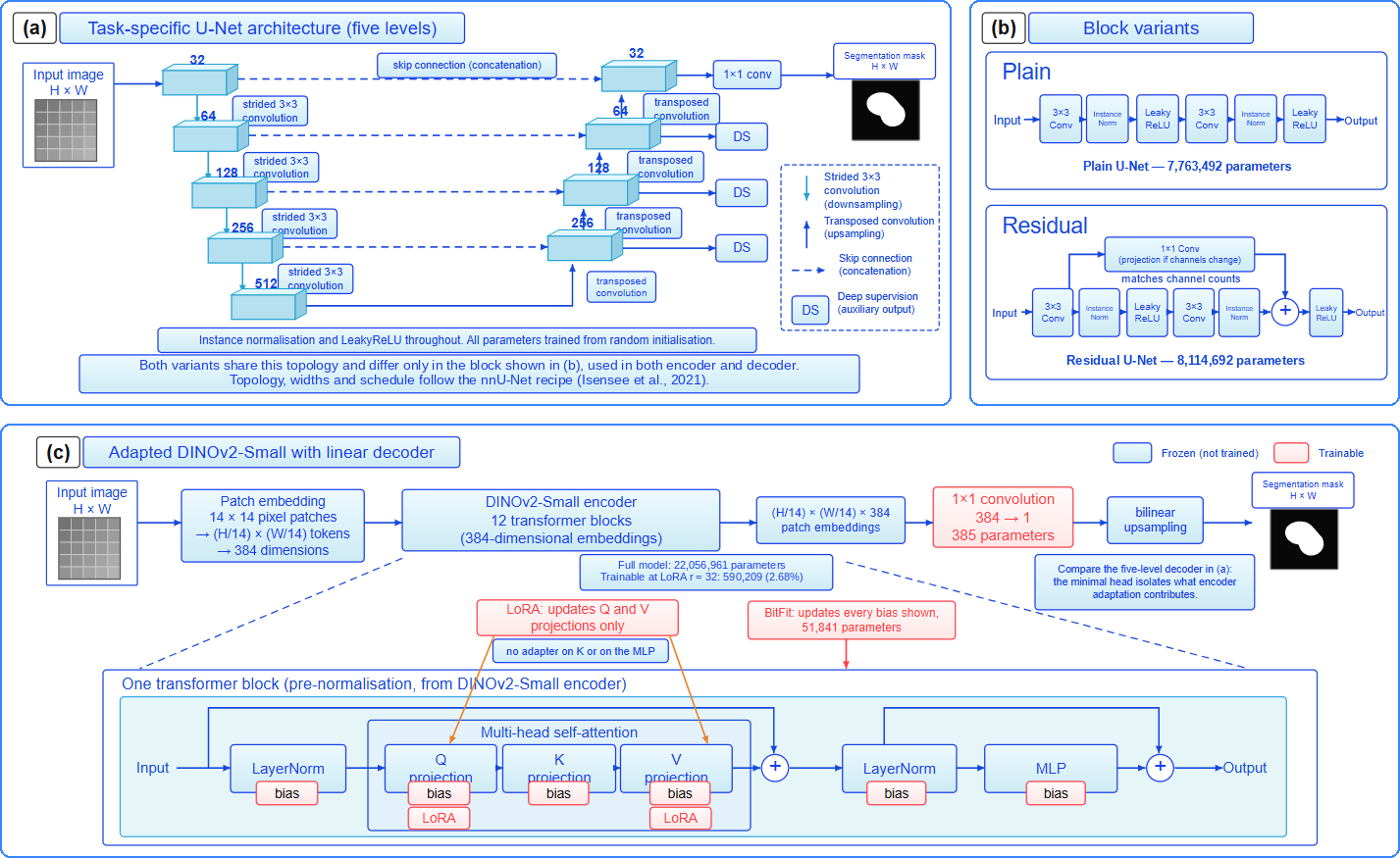}
\caption{Model architectures. (a) The five-level encoder--decoder topology shared by both task-specific comparators. (b) One block in each variant. The two differ in the residual addition and its projection shortcut, applied in both encoder and decoder, while widths, normalisation, activation, downsampling and training recipe are identical. (c) The adapted model: a DINOv2-Small encoder producing 384-dimensional patch embeddings, reduced to a one-channel logit map by a single $1{\times}1$ convolution of 385 parameters and upsampled to the input resolution. The magnified block shows where each adaptation strategy intervenes. Blue marks parameters held frozen and red marks parameters updated.}
\label{fig:arch}
\end{figure*}

All adaptation experiments use a DINOv2-Small encoder \citep{oquab2024dinov2}, the smallest of the released variants at 22.1 million parameters against 1.1 billion for the largest, with a linear decoder head: a single $1{\times}1$ convolution mapping the 384-dimensional patch embeddings to a one-channel logit map, bilinearly upsampled to the input resolution, containing 385 parameters (Fig.~\ref{fig:arch}c). This minimality is intentional: a heavier decoder would absorb part of the adaptation and confound the question of how much the encoder must change. Two consequences bound what Section~\ref{sec:results} shows. The frozen configuration should be read as a minimal baseline; \citet{li2025dinounet} combine a frozen DINOv3 backbone with an adapter, a projection module and a U-Net-style decoder, a substantially different architecture. And the adapted model is not intended to compete with architectures developed for a specific task, which pair large encoders with multi-scale decoders of their own; Section~\ref{sec:arch} states that scope where published scores are introduced.

Seven configurations span the capacity axis (Table~\ref{tab:strategies}); their results by test set are given in Table~\ref{tab:capacity}. The frozen configuration trains only the decoder and establishes how much of the task the pretrained representation already solves. BitFit \citep{benzaken2022bitfit} trains all encoder bias terms with the decoder; having no rank parameter and intervening throughout the encoder, it probes the capacity axis independently of LoRA's mechanism, which is its role in Section~\ref{sec:capacity}. LoRA is applied at ranks 4, 8, 16 and 32, with low-rank updates attached to the query and value projections of every attention block, the placement recommended by \citet{hu2022lora}, and scaling $\alpha = 2r$, holding $\alpha/r$ fixed so that the sweep varies rank alone. Dropout is 0.1, and rank-stabilised scaling \citep{kalajdzievski2023rslora} was disabled for the same reason. Full fine-tuning updates all encoder weights together with the decoder.

\begin{table}[tp]
\centering
\caption{Adaptation strategies on the capacity axis. Every configuration also trains the 385-parameter linear decoder, which is included in the trainable count; the percentage is relative to full fine-tuning. Q and V: query and value projections of every attention block.}
\label{tab:strategies}
\footnotesize
\begin{tabular*}{\columnwidth}{@{\extracolsep{\fill}}lrrl@{}}
\toprule
Configuration & Trainable & \% of full FT & Encoder parameters updated \\
\midrule
Frozen & 385 & 0.002 & None \\
BitFit & 51,841 & 0.24 & All bias terms \\
LoRA $r{=}4$ & 74,113 & 0.34 & Rank-4 factors on Q and V \\
LoRA $r{=}8$ & 147,841 & 0.67 & Rank-8 factors on Q and V \\
LoRA $r{=}16$ & 295,297 & 1.34 & Rank-16 factors on Q and V \\
LoRA $r{=}32$ & 590,209 & 2.68 & Rank-32 factors on Q and V \\
Full FT & 22,056,961 & 100 & All weights \\
\bottomrule
\end{tabular*}
\end{table}

All strategies use AdamW with a two-epoch warmup, three for full fine-tuning, cosine decay, weight decay 1e-4 and early stopping on validation DSC. The optimiser, the schedule and the stopping rule are held identical across strategies by design, since the adaptation axis is intended to vary one thing, where and how many parameters are trained, and a schedule that moved with the strategy would confound that variation with it. The learning rate is the exception, for the reason given below. \citet{zhang2023samed} report that warmup stabilises SAMed fine-tuning and that AdamW further improves performance. The frozen strategy trains its decoder at 1e-3, LoRA uses 1e-4 throughout, and full fine-tuning uses layer-wise decay \citep{howard2018ulmfit} with a base encoder rate of 5e-6 decaying by 0.75 per layer, a decay within the range conventionally used for vision transformer fine-tuning. BitFit uses one optimiser group at 1e-3, 3e-3 and 1e-3 for polyp, breast and skin, selected over \{1e-2, 3e-3, 1e-3, 3e-4\} on in-domain validation.

The governing principle is that each strategy is evaluated at its own selected setting rather than a shared one, and this is more consequential than it appears. An earlier run of this study trained full fine-tuning at a single flat encoder rate for a fixed fifteen epochs, without layer-wise decay or warmup, and under that configuration it appeared clearly inferior, a conclusion that would have been an artefact of the protocol rather than a property of the methods. One asymmetry follows: BitFit received a per-domain learning-rate search whereas LoRA and full fine-tuning did not, which favours BitFit, and Section~\ref{sec:capacity} should be read accordingly.

\subsection{Task-specific architectures}\label{sec:tasksp}

A comparison against one task-specific model leaves an evident objection: any deficit might reflect a weakness in that implementation rather than a property of task-specific training. Two architectures are therefore used, bracketing the task-specific axis (Fig.~\ref{fig:arch}a, b). The residual U-Net represents the residual design family and the plain U-Net the field's conventional reference point, and if both fall behind the deficit cannot be attributed to a single architectural choice.

\subsubsection{Residual U-Net}
The residual variant is in the spirit of nnU-Net ResEnc \citep{isensee2024revisited}, with residual blocks in both encoder and decoder, trained under the nnU-Net recipe \citep{isensee2021nnunet}: 8,114,692 parameters in a five-level encoder of base width 32 with residual blocks, instance normalisation, LeakyReLU and deep supervision at every decoder level. The original U-Net of \citet{ronneberger2015unet} was deliberately not adopted, since using the 2015 configuration would have reproduced the weak-baseline error in the direction that flatters this study.

\subsubsection{Plain U-Net}
The plain variant is the plain convolutional U-Net that nnU-Net configures by default \citep{isensee2021nnunet, isensee2024revisited}, identical except that the residual addition and $1{\times}1$ skip projection are removed, at 7,763,492 parameters.

\subsubsection{Shared training recipe}
Both follow the nnU-Net recipe \citep{isensee2021nnunet}: stochastic gradient descent with momentum 0.99 and Nesterov acceleration, a polynomial schedule, weight decay 3e-5, aggressive augmentation, deep supervision and a combined Dice and cross-entropy objective. The augmentation policy was determined empirically, and that search is reported in Section~\ref{sec:datasize} where it also tests whether the deficit is a configuration failure. Learning rates were searched independently for each architecture on polyp over an identical grid, selecting 3e-3 for the residual variant and 1e-3 for the plain one. In the skin domain both architectures ran at 3e-3; the plain variant's polyp-selected rate was not additionally evaluated there. One consequence bears on Section~\ref{sec:arch}: the skin comparison is the more controlled of the three, both models operating at the same rate, whereas in polyp and breast each is at its own selected setting. The comparison is therefore two models each at its own best setting rather than a controlled ablation, a limitation noted in Section~\ref{sec:limitations} along with the search having been conducted on polyp alone.

\subsection{Training protocol and model selection}\label{sec:training}

DINOv2 operates on fourteen-pixel patches, so resolutions are multiples of fourteen: $224{\times}224$ at batch size 16 for polyp and breast, and $392{\times}392$ at batch size 8 for skin, where dermoscopy boundaries are fine relative to the field of view. The task-specific models in skin run at $384{\times}384$, because 392 is not divisible by the cumulative downsampling factor of their encoder. The data-size sweep of Section~\ref{sec:datasize} was run at 224 across all skin fractions to keep cost tractable, so those figures sit at a different operating point and are compared only against themselves.

Adaptation strategies converge within tens of epochs, whereas task-specific models require an order of magnitude more; within any single comparison the budget is identical, so no configuration is advantaged by training duration. Task-specific models were trained for 300 epochs in polyp and breast and 100 in skin (Table~\ref{tab:budgets}). Adaptation ran at two budgets, 20 epochs for the core comparison and 40 for the capacity analyses following the budget control of Section~\ref{sec:capacity}, and figures from the two are never combined within a domain. In the data-size sweep, budgets were scaled inversely with subset size so each fraction received approximately equal optimiser steps, without which the deficit would confound data volume with training duration.

\begin{table}[tbp]
\centering
\caption{Epoch budgets, matched within every comparison but not across domains.}
\label{tab:budgets}
\footnotesize
\setlength{\tabcolsep}{3pt}
\begin{tabular*}{\columnwidth}{@{\extracolsep{\fill}}lccc@{}}
\toprule
Configuration & Polyp & Breast & Skin \\
\midrule
Adaptation, capacity axis & 40 & 40 & 20 \\
Adaptation, core multi-seed comparison & 20 & -- & 20 \\
From-scratch & 300 & 300 & 100 \\
Data-size sweep & \multicolumn{3}{c}{scaled inversely with subset size} \\
\bottomrule
\end{tabular*}
\end{table}

The programme these budgets describe is substantial. Each of the three domains carries seven adaptation configurations on the capacity axis, two task-specific architectures, three or four training-set fractions, and a learning-rate search over four values for BitFit, with the same search run in polyp for each of the two task-specific models; the core comparison between LoRA and full fine-tuning is run at three seeds and at two ranks, and the capacity axis at two budgets. Running few methods at many operating points, which Section~\ref{sec:rw-gap} identifies as the trade-off this study makes, is what accounts for that count, and it is also what makes Section~\ref{sec:stability} possible.

Checkpoints, learning rates, ranks and configurations were all selected on in-domain validation DSC, and no out-of-distribution test set influenced any selection decision at any stage. This is emphasised because it is load-bearing: a recurring observation is that the configuration optimal in domain is not the one optimal out of distribution, and that carries weight only if selection was genuinely blind to those sets.

\subsection{Reproducibility, the measurement floor and evaluation}\label{sec:eval}

All experiments used PyTorch \citep{paszke2019pytorch}, HuggingFace Transformers \citep{wolf2020transformers} and the PEFT library, with mixed precision \citep{micikevicius2018mixed}, on NVIDIA L4 and A100 GPUs, with checkpoints written to persistent storage so that interrupted sessions could resume without retraining. Seed 42 was used throughout, with 123 and 2024 added for the core comparison of LoRA against full fine-tuning. BitFit and the plain U-Net were run at seed 42 only and are reported as single-seed signal rather than multi-seed evidence, and the two terms are used with that distinction throughout.

Seeding does not make these runs deterministic. Seeding covered the main process only, leaving the dataloader workers' augmentation sequence non-deterministic, and cuDNN determinism was not enforced, so two runs at an identical seed may diverge. This was observed directly: a repeated BitFit polyp run differed by up to 0.013 DSC out of distribution and 0.004 on validation. Determinism was not enforced partway through because that would have made later runs non-comparable with earlier ones. Two skin analyses initially contained an evaluation error; both were re-evaluated from the same checkpoints, and all figures reported here are the corrected ones.

That variation is adopted as a measurement floor, and its scope requires precise statement because it does not apply uniformly. The band was derived from the difference between two single runs and therefore bounds point comparisons: where a configuration is represented by one number, a difference below 0.013 DSC cannot be distinguished from run-to-run variation and is not interpreted, irrespective of sign. This governs the capacity axis, both task-specific architectures, the data-size sweep and every learning-rate search. It is a reference drawn from one repeated configuration rather than a measured bound for every experiment.

It does not govern the paired significance tests of Section~\ref{sec:capacity}, which measure a different quantity. The Wilcoxon signed-rank test applied to seed-averaged per-image vectors assesses not the magnitude of a difference but its consistency across images: a small difference holding in the same direction on most images is significant, while a larger one varying erratically is not. The polyp comparison demonstrates this within a single table, where CVC-ColonDB shows a difference of $+0.0022$ with the lowest $p$-value of the five at 0.0001, while ETIS shows the largest at $+0.0130$ and is not significant at $p = 0.1191$. Reporting either criterion alone would misdescribe the comparison, so multi-seed results are reported by direction and $p$-value with magnitude given but not used as a threshold, and single-run comparisons against the band and never with a $p$-value. Because the signed-rank statistic is computed on the ranks of the absolute per-image differences, magnitude enters the test through those ranks; and because images rather than seeds are the unit of pairing, the test does not directly assess variation across seeds.

Segmentation quality is reported with six metrics: the Dice similarity coefficient (DSC) and intersection over union (IoU) for overlap; sensitivity and precision, which separate a missed lesion from over-segmentation; and, for the boundary, two spatial-distance metrics in the taxonomy of \citet{taha2015metrics}, the 95th-percentile Hausdorff distance (HD95), taken over the boundary distances in both directions, and the average symmetric surface distance (ASSD), the mean of the two directed mean boundary distances, together with the normalised surface Dice (NSD) \citep{nikolov2021nsd}, the fraction of boundary points of each mask lying within a tolerance of the other's boundary. All are accumulated per image rather than pooled, since pooling weights each image by its lesion area; per-image accumulation also yields a vector rather than a scalar, which is what makes significance testing possible. Where multiple seeds are available, comparisons are tested by Wilcoxon signed-rank \citep{wilcoxon1945} per test set, preferred to a paired $t$-test because per-image DSC distributions are strongly left-skewed, and to a Mann--Whitney test because it retains the pairing. Single-seed results are not tested, since a $p$-value on one run would characterise variation across images rather than across runs.

Distances are measured in pixels at the resolution each model was evaluated at, because most of the datasets do not record physical pixel spacing. That resolution differs between domains, and within skin between the adapted and the task-specific models (Section~\ref{sec:training}), so HD95 and ASSD are also reported divided by the image diagonal; only the normalised values are compared across resolutions, and pixel values only within one. The NSD tolerance is defined as a fixed fraction of the image width, 2/224, which is 2 pixels at 224, 3.4 at 384 and 3.5 at 392; results at twice that tolerance are given in Table~\ref{tab:boundary}. Where exactly one of the two masks is empty, HD95 and ASSD are set to the image diagonal at that resolution and NSD to zero, and the number of such images is reported together with the boundary metrics over the images where both masks are non-empty. A prediction that is empty when a lesion is present is counted as precision zero. No reference mask was empty at the evaluated resolution in any test set, so the case of an empty reference never arose.

DSC was the only criterion used for selection. The other five metrics were computed afterwards from the same checkpoints, and no model or configuration choice depends on them. Before they were computed, the per-image DSC of every checkpoint was recomputed and compared with the corresponding values reported in this paper, 130 checks in all across the tables of this paper; every value agreed to within 0.001, the largest deviation being 0.0007.

\section{Results}\label{sec:results}

\subsection{In-domain parity, out-of-distribution deficit}\label{sec:main}

This section reports the direct comparison, using the strongest configuration of each approach: DINOv2 with LoRA at rank 32, selected on in-domain validation, against the two task-specific architectures (Table~\ref{tab:main}). DSC is discussed first, as the metric every selection decision used; the other five metrics follow.

\begin{table*}[tp]
\centering
\caption{Six metrics by model and test set. DSC, IoU, sensitivity (Sens), precision (Prec) and NSD: higher is better; HD95 and ASSD are divided by the image diagonal (Section~\ref{sec:eval}): lower is better. NSD at a tolerance of 2/224 of the image width. $\varnothing$: images with an empty prediction. Resolution 224\,px in polyp and breast; in skin 392\,px for DINOv2 and 384\,px for the U-Nets. Task-specific models trained for 300 epochs in polyp and breast and 100 in skin; DINOv2 for 40 epochs in polyp and breast and 20 in skin, single-seed except skin DINOv2, a three-seed mean (its $\varnothing$ is the mean per seed). (in) marks in-domain sets; BUSI-val also served for model selection. Asterisks mark skin sets drawn from the same archive as the training data; all CVC-T images also occur in CVC-ColonDB (Section~\ref{sec:datasets}). DSC values are those of the original evaluation, which the re-evaluation that produced the other metrics reproduces to within 0.0007 (Section~\ref{sec:eval}). Bold marks the best value for each test set.}
\label{tab:main}
\footnotesize
\setlength{\tabcolsep}{1.5pt}
\begin{tabular*}{\textwidth}{@{\extracolsep{\fill}}lrlcccccccr@{}}
\toprule
Test set & $n$ & Model & DSC & IoU & Sens & Prec & HD95 & ASSD & NSD & $\varnothing$ \\
\midrule
\multicolumn{11}{@{}l}{\textit{Polyp}} \\
Kvasir (in) & 100 & Plain U-Net & 0.8804 & 0.813 & 0.875 & 0.917 & 0.071 & 0.017 & \textbf{0.682} & 0 \\
 &  & ResU-Net & 0.8611 & 0.793 & 0.858 & 0.907 & 0.067 & 0.018 & 0.669 & 0 \\
 &  & LoRA $r{=}32$ & \textbf{0.8877} & \textbf{0.823} & \textbf{0.883} & \textbf{0.918} & \textbf{0.058} & \textbf{0.016} & 0.649 & 0 \\
\addlinespace[2pt]
CVC-ClinicDB (in) & 62 & Plain U-Net & 0.8803 & 0.831 & 0.892 & 0.874 & \textbf{0.042} & \textbf{0.016} & 0.782 & 0 \\
 &  & ResU-Net & \textbf{0.8826} & \textbf{0.833} & 0.886 & \textbf{0.889} & 0.047 & 0.027 & \textbf{0.791} & 1 \\
 &  & LoRA $r{=}32$ & 0.8684 & 0.798 & \textbf{0.892} & 0.871 & 0.046 & 0.026 & 0.653 & 1 \\
\addlinespace[2pt]
CVC-ColonDB & 380 & Plain U-Net & 0.7382 & 0.646 & 0.812 & 0.744 & 0.120 & \textbf{0.037} & \textbf{0.538} & 0 \\
 &  & ResU-Net & 0.7240 & 0.635 & 0.763 & 0.751 & 0.106 & 0.040 & 0.532 & 1 \\
 &  & LoRA $r{=}32$ & \textbf{0.7694} & \textbf{0.683} & \textbf{0.816} & \textbf{0.780} & \textbf{0.086} & 0.049 & 0.515 & 10 \\
\addlinespace[2pt]
ETIS & 196 & Plain U-Net & 0.6321 & 0.552 & 0.779 & 0.588 & 0.174 & 0.066 & \textbf{0.496} & 0 \\
 &  & ResU-Net & 0.6179 & 0.534 & 0.765 & 0.568 & 0.184 & 0.078 & 0.471 & 2 \\
 &  & LoRA $r{=}32$ & \textbf{0.6984} & \textbf{0.609} & \textbf{0.805} & \textbf{0.654} & \textbf{0.099} & \textbf{0.041} & 0.478 & 2 \\
\addlinespace[2pt]
CVC-T & 60 & Plain U-Net & 0.8406 & 0.754 & 0.948 & 0.788 & 0.043 & 0.013 & 0.652 & 0 \\
 &  & ResU-Net & 0.7941 & 0.716 & 0.877 & 0.750 & 0.053 & 0.018 & 0.628 & 0 \\
 &  & LoRA $r{=}32$ & \textbf{0.8665} & \textbf{0.792} & \textbf{0.951} & \textbf{0.827} & \textbf{0.032} & \textbf{0.010} & \textbf{0.672} & 0 \\
\midrule
\multicolumn{11}{@{}l}{\textit{Breast}} \\
BUSI-val (in) & 54 & Plain U-Net & 0.8177 & 0.732 & 0.835 & 0.826 & \textbf{0.063} & \textbf{0.022} & 0.551 & 0 \\
 &  & ResU-Net & \textbf{0.8244} & \textbf{0.742} & \textbf{0.845} & \textbf{0.830} & 0.075 & 0.024 & \textbf{0.571} & 0 \\
 &  & LoRA $r{=}32$ & 0.8029 & 0.709 & 0.844 & 0.788 & 0.072 & 0.028 & 0.438 & 0 \\
\addlinespace[2pt]
BrEaST & 252 & Plain U-Net & 0.6545 & 0.550 & 0.757 & 0.663 & 0.124 & 0.037 & 0.367 & 0 \\
 &  & ResU-Net & 0.6463 & 0.542 & 0.766 & 0.638 & 0.136 & 0.039 & \textbf{0.378} & 0 \\
 &  & LoRA $r{=}32$ & \textbf{0.7133} & \textbf{0.604} & \textbf{0.813} & \textbf{0.711} & \textbf{0.075} & \textbf{0.029} & 0.338 & 1 \\
\addlinespace[2pt]
BUS-UCLM & 260 & Plain U-Net & 0.6320 & 0.532 & 0.745 & 0.618 & 0.161 & 0.048 & \textbf{0.373} & 0 \\
 &  & ResU-Net & 0.5981 & 0.496 & 0.725 & 0.569 & 0.189 & 0.055 & 0.340 & 0 \\
 &  & LoRA $r{=}32$ & \textbf{0.7046} & \textbf{0.600} & \textbf{0.783} & \textbf{0.738} & \textbf{0.082} & \textbf{0.031} & 0.349 & 1 \\
\midrule
\multicolumn{11}{@{}l}{\textit{Skin}} \\
ISIC 2016* & 269 & Plain U-Net & 0.9123 & 0.847 & 0.917 & \textbf{0.926} & 0.039 & 0.014 & 0.430 & 0 \\
 &  & ResU-Net & 0.9084 & 0.842 & 0.930 & 0.909 & 0.042 & 0.015 & 0.415 & 0 \\
 &  & LoRA $r{=}32$ & \textbf{0.924} & \textbf{0.864} & \textbf{0.939} & 0.923 & \textbf{0.029} & \textbf{0.012} & \textbf{0.431} & 0 \\
\addlinespace[2pt]
ISIC 2017* & 172 & Plain U-Net & 0.8336 & 0.746 & 0.810 & 0.919 & 0.065 & 0.025 & 0.393 & 0 \\
 &  & ResU-Net & 0.8356 & 0.748 & 0.823 & 0.907 & 0.068 & 0.025 & 0.384 & 0 \\
 &  & LoRA $r{=}32$ & \textbf{0.872} & \textbf{0.793} & \textbf{0.851} & \textbf{0.939} & \textbf{0.047} & \textbf{0.020} & \textbf{0.399} & 0 \\
\addlinespace[2pt]
PH2 & 200 & Plain U-Net & 0.9167 & 0.851 & 0.972 & 0.876 & 0.041 & 0.016 & 0.341 & 0 \\
 &  & ResU-Net & 0.9032 & 0.831 & 0.975 & 0.852 & 0.048 & 0.018 & 0.308 & 0 \\
 &  & LoRA $r{=}32$ & \textbf{0.925} & \textbf{0.865} & \textbf{0.985} & \textbf{0.879} & \textbf{0.031} & \textbf{0.013} & \textbf{0.386} & 0 \\
\bottomrule
\end{tabular*}
\end{table*}

\begin{figure*}[t]
\centering
\includegraphics[width=\linewidth]{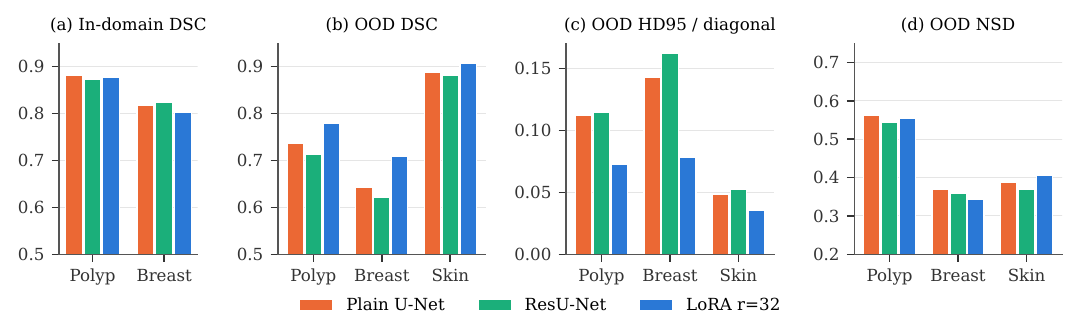}
\caption{Mean performance by domain, model and evaluation regime, averaged over each domain's test sets in that regime. (a) In-domain DSC; (b)--(d) out-of-distribution DSC, HD95 divided by the image diagonal (lower is better) and NSD at 2/224 of the image width. Skin has no in-domain test set. The ordering between the two model families reverses between (a) and (b); (c) follows (b), while (d) does not.}
\label{fig:regimes}
\end{figure*}

The two regimes give opposite answers. In domain, adaptation confers no advantage. Averaged over the two in-domain polyp sets the plain U-Net reaches 0.8803 against 0.8780, a difference well below the 0.013 measurement floor (Section~\ref{sec:eval}), and the residual variant is behind by $-0.006$, likewise within it; in breast both task-specific models are ahead on a validation split that also served for early stopping, so this result is indicative rather than conclusive. Within the distribution a model was trained on, a conventional U-Net trained from random initialisation therefore matches an adapted DINOv2, at roughly a third of the parameters and, as Section~\ref{sec:datasize} shows, in breast with as few as 119 training images.

Out of distribution the picture reverses on all eight sets, under both architectures, without exception (Fig.~\ref{fig:regimes}; per-set deficits in Table~\ref{tab:deficit}). But the uniformity of the sign should not obscure the variation in magnitude: the sixteen deficits range from $-0.008$ (PH2, plain U-Net) to $-0.107$ (BUS-UCLM, ResU-Net), and two fall within the floor, both in skin. Skin is therefore not an exception in direction but close to one in magnitude. Two considerations bear on this: it has by far the largest training set, and two of its three test sets originate from the same archive as the training data, representing a milder shift. PH2, the one genuinely independent skin source, shows the smallest deficit of the three under the plain U-Net, which does not support the second explanation. Section~\ref{sec:datasize} addresses the first.

\begin{table*}[tbp]
\centering
\begin{minipage}[t]{\pairwidth}
\centering
\caption{Deficit of each task-specific model relative to DINOv2 + LoRA $r{=}32$ on the eight out-of-distribution sets. Negative values indicate the task-specific model is behind; a dagger marks differences below the 0.013 floor.}
\label{tab:deficit}
\footnotesize
\setlength{\tabcolsep}{3pt}
\begin{tabular*}{\linewidth}{@{\extracolsep{\fill}}llcc@{}}
\toprule
Domain & Test set & Plain U-Net & ResU-Net \\
\midrule
Polyp  & CVC-ColonDB & $-0.031$ & $-0.045$ \\
Polyp  & ETIS        & $-0.066$ & $-0.081$ \\
Polyp  & CVC-T       & $-0.026$ & $-0.072$ \\
\midrule
Breast & BrEaST      & $-0.059$ & $-0.067$ \\
Breast & BUS-UCLM    & $-0.073$ & $-0.107$ \\
\midrule
Skin   & ISIC 2016*  & $-0.012$$^{\dagger}$ & $-0.016$ \\
Skin   & ISIC 2017*  & $-0.038$ & $-0.036$ \\
Skin   & PH2         & $-0.008$$^{\dagger}$ & $-0.022$ \\
\bottomrule
\end{tabular*}
\end{minipage}\pairsep
\begin{minipage}[t]{\pairwidth}
\centering
\caption{Plain U-Net minus ResU-Net by test set, all domains, single seed. Positive values favour the plain architecture; a dagger marks differences below the floor.}
\label{tab:plainres}
\footnotesize
\setlength{\tabcolsep}{3pt}
\begin{tabular*}{\linewidth}{@{\extracolsep{\fill}}llc@{}}
\toprule
Domain & Test set & Difference \\
\midrule
Polyp  & Kvasir (in-domain)       & $+0.019$ \\
Polyp  & CVC-ClinicDB (in-domain) & $-0.002$$^{\dagger}$ \\
Polyp  & CVC-ColonDB              & $+0.014$ \\
Polyp  & ETIS                     & $+0.014$ \\
Polyp  & CVC-T                    & $+0.047$ \\
\midrule
Breast & BUSI-val (in-domain)     & $-0.007$$^{\dagger}$ \\
Breast & BrEaST                   & $+0.008$$^{\dagger}$ \\
Breast & BUS-UCLM                 & $+0.034$ \\
\midrule
Skin   & ISIC 2016*               & $+0.004$$^{\dagger}$ \\
Skin   & ISIC 2017*               & $-0.002$$^{\dagger}$ \\
Skin   & PH2                      & $+0.014$ \\
\bottomrule
\end{tabular*}
\end{minipage}
\end{table*}

\begin{figure*}[t]
\centering
\includegraphics[width=\linewidth]{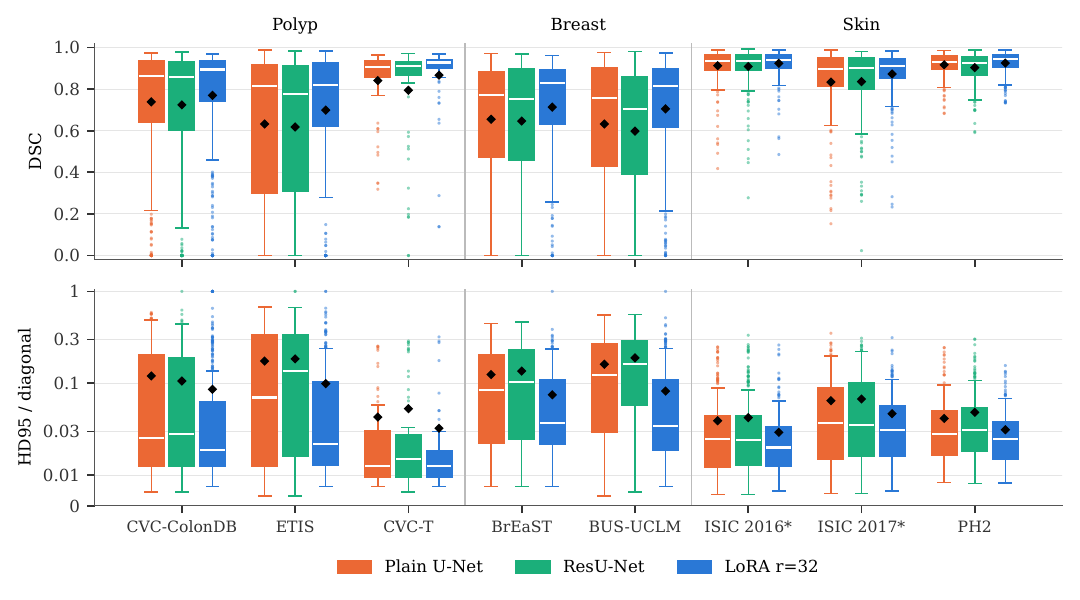}
\caption{Per-image DSC (top) and HD95 divided by the image diagonal (bottom, symmetric-log scale; lower is better) on the eight out-of-distribution sets. Boxes span the interquartile range with the median in white; whiskers extend to 1.5 times the interquartile range; diamonds mark the mean. An empty prediction scores DSC 0 and HD95 1. Models and budgets as in Table~\ref{tab:main}; skin DINOv2 values are per-image means over three seeds.}
\label{fig:dist}
\end{figure*}

The other metrics agree with DSC out of distribution except at the finest scale of the boundary. DINOv2 is best on all eight out-of-distribution sets in IoU, sensitivity and HD95, and on seven in precision and ASSD; the exceptions are precision on ISIC 2016*, and ASSD on CVC-ColonDB, where the maximum distance assigned to its empty predictions (Section~\ref{sec:failure}) outweighs its advantage on the remaining images. NSD at the 2/224 tolerance does not follow: a task-specific model leads on four of the eight sets, all four evaluated at 224 pixels, and in domain both task-specific models lead DINOv2 on all three in-domain sets, by up to 0.14 on CVC-ClinicDB. Fig.~\ref{fig:regimes} summarises these quantities by domain and regime, and Fig.~\ref{fig:dist} shows the per-image distributions behind the out-of-distribution means. The advantage is a shift of the distribution rather than a uniform gain: DINOv2 has the higher DSC on between 55 and 62 per cent of out-of-distribution images, depending on domain and comparator, and every model retains a long lower tail on CVC-ColonDB, ETIS and both breast sets. Distances in pixels, NSD at the secondary tolerance and the boundary metrics over non-empty predictions are given in Table~\ref{tab:boundary}.

\begin{table*}[p]
\centering
\caption{Boundary metrics for the models of Table~\ref{tab:main}, in the units not shown there. HD95 and ASSD in pixels at the evaluated resolution (224 px in polyp and breast; skin 392 px for DINOv2, 384 px for the U-Nets), so pixel values are comparable only within a resolution; NSD at the secondary tolerance of 4/224 of the image width. The last three columns repeat HD95, ASSD (divided by the image diagonal) and NSD at 2/224 over the images on which the model's prediction is not empty, so that the penalty assigned to empty predictions is visible as the difference from Table~\ref{tab:main}. Skin DINOv2 values are three-seed means of per-image values; an image counts as empty for it when its seed-averaged empty indicator is at least 0.5.}
\label{tab:boundary}
\footnotesize
\setlength{\tabcolsep}{1.5pt}
\begin{tabular*}{\textwidth}{@{\extracolsep{\fill}}lllcccccc@{}}
\toprule
 & & & \multicolumn{3}{c}{All images} & \multicolumn{3}{c}{Non-empty predictions} \\
\cmidrule(lr){4-6}\cmidrule(lr){7-9}
Domain & Test set & Model & HD95 (px) & ASSD (px) & NSD$_4$ & HD95 & ASSD & NSD$_2$ \\
\midrule
Polyp & Kvasir & Plain U-Net & 22.6 & 5.5 & 0.794 & 0.071 & 0.017 & 0.682 \\
 &  & ResU-Net & 21.3 & 5.8 & 0.776 & 0.067 & 0.018 & 0.669 \\
 &  & LoRA $r{=}32$ & 18.4 & 5.2 & 0.821 & 0.058 & 0.016 & 0.649 \\
\addlinespace[2pt]
 & CVC-ClinicDB & Plain U-Net & 13.2 & 5.1 & 0.868 & 0.042 & 0.016 & 0.782 \\
 &  & ResU-Net & 14.9 & 8.5 & 0.881 & 0.031 & 0.011 & 0.804 \\
 &  & LoRA $r{=}32$ & 14.6 & 8.1 & 0.838 & 0.031 & 0.010 & 0.663 \\
\addlinespace[2pt]
 & CVC-ColonDB & Plain U-Net & 38.0 & 11.8 & 0.688 & 0.120 & 0.037 & 0.538 \\
 &  & ResU-Net & 33.5 & 12.5 & 0.681 & 0.103 & 0.037 & 0.533 \\
 &  & LoRA $r{=}32$ & 27.3 & 15.5 & 0.719 & 0.061 & 0.023 & 0.529 \\
\addlinespace[2pt]
 & ETIS & Plain U-Net & 55.2 & 20.9 & 0.593 & 0.174 & 0.066 & 0.496 \\
 &  & ResU-Net & 58.4 & 24.6 & 0.574 & 0.176 & 0.068 & 0.476 \\
 &  & LoRA $r{=}32$ & 31.4 & 13.1 & 0.669 & 0.090 & 0.032 & 0.483 \\
\addlinespace[2pt]
 & CVC-T & Plain U-Net & 13.6 & 4.0 & 0.835 & 0.043 & 0.013 & 0.652 \\
 &  & ResU-Net & 16.8 & 5.8 & 0.800 & 0.053 & 0.018 & 0.628 \\
 &  & LoRA $r{=}32$ & 10.3 & 3.2 & 0.871 & 0.032 & 0.010 & 0.672 \\
\midrule
Breast & BUSI-val & Plain U-Net & 19.9 & 6.8 & 0.703 & 0.063 & 0.022 & 0.551 \\
 &  & ResU-Net & 23.7 & 7.6 & 0.722 & 0.075 & 0.024 & 0.571 \\
 &  & LoRA $r{=}32$ & 22.8 & 8.9 & 0.661 & 0.072 & 0.028 & 0.438 \\
\addlinespace[2pt]
 & BrEaST & Plain U-Net & 39.4 & 11.8 & 0.533 & 0.124 & 0.037 & 0.367 \\
 &  & ResU-Net & 43.1 & 12.5 & 0.536 & 0.136 & 0.039 & 0.378 \\
 &  & LoRA $r{=}32$ & 23.8 & 9.2 & 0.541 & 0.071 & 0.025 & 0.339 \\
\addlinespace[2pt]
 & BUS-UCLM & Plain U-Net & 51.0 & 15.3 & 0.500 & 0.161 & 0.048 & 0.373 \\
 &  & ResU-Net & 59.8 & 17.5 & 0.454 & 0.189 & 0.055 & 0.340 \\
 &  & LoRA $r{=}32$ & 26.1 & 9.7 & 0.554 & 0.079 & 0.027 & 0.350 \\
\midrule
Skin & ISIC 2016* & Plain U-Net & 21.3 & 7.8 & 0.656 & 0.039 & 0.014 & 0.430 \\
 &  & ResU-Net & 23.0 & 8.4 & 0.644 & 0.042 & 0.015 & 0.415 \\
 &  & LoRA $r{=}32$ & 16.2 & 6.6 & 0.698 & 0.029 & 0.012 & 0.431 \\
\addlinespace[2pt]
 & ISIC 2017* & Plain U-Net & 35.1 & 13.5 & 0.555 & 0.065 & 0.025 & 0.393 \\
 &  & ResU-Net & 36.7 & 13.7 & 0.553 & 0.068 & 0.025 & 0.384 \\
 &  & LoRA $r{=}32$ & 25.9 & 10.8 & 0.599 & 0.047 & 0.020 & 0.399 \\
\addlinespace[2pt]
 & PH2 & Plain U-Net & 22.5 & 8.5 & 0.588 & 0.041 & 0.016 & 0.341 \\
 &  & ResU-Net & 26.3 & 10.0 & 0.531 & 0.048 & 0.018 & 0.308 \\
 &  & LoRA $r{=}32$ & 17.3 & 7.4 & 0.636 & 0.031 & 0.013 & 0.386 \\
\bottomrule
\end{tabular*}
\end{table*}

\subsection{The deficit is not architectural}\label{sec:arch}

Section~\ref{sec:main} rests on an assumption not yet examined: that the task-specific models are competently configured. If they are not, the deficits measure the quality of the implementation rather than any property of task-specific training.

The two architectures agree. They differ architecturally only in the residual connection, applied in both encoder and decoder, and are otherwise identical in widths, downsampling, normalisation, activation, deep supervision and recipe, each at its own selected learning rate. The deficit is negative in all sixteen comparisons: two models at different points on the task-specific design space, tuned independently, place the same eight sets on the same side.

A secondary result runs against expectation. The plain U-Net is ahead of the residual variant on seven of the eight out-of-distribution sets and behind on one, ISIC 2017* by 0.002, and five differences exceed the floor (Table~\ref{tab:plainres}). Under the conditions here, two-dimensional and binary with 479 to 2,075 training images, the residual variant trained under the nnU-Net recipe therefore did not outperform its plain counterpart, and out of distribution it is the weaker of the two. Two qualifications apply: the comparison is single-seed, so it rests on the five differences clearing the floor, and in polyp and breast the architectures ran at different learning rates, so part of the difference there may originate in the optimiser rather than the residual connection. Skin is the exception and is therefore the most informative of the three: both architectures ran at 3e-3, making that comparison a single-variable ablation, and the plain variant is ahead on two of its three test sets. What the comparison establishes is that no reasonable task-specific configuration closes the gap.

The internal comparison shows the deficit does not depend on the architecture chosen; it does not show either is strong in absolute terms. Because polyp follows the protocol of \citet{fan2020pranet}, published scores are directly comparable in setup (Table~\ref{tab:published}).

\begin{table}[t]
\centering
\caption{Comparison with published results (DSC) on the three out-of-distribution polyp sets under the PraNet protocol \citep{fan2020pranet}. Rows from this work are single-seed. U-Net, U-Net++, ACSNet, MSEG and Polyp-PVT scores are as reported by \citet{dong2023polyppvt}; SAM-Mamba scores are from \citet{dutta2025sammamba}. Citations in the model column identify the original methods.}
\label{tab:published}
\footnotesize
\begin{tabular*}{\columnwidth}{@{\extracolsep{\fill}}lccc@{}}
\toprule
Model & CVC-ColonDB & ETIS & CVC-T \\
\midrule
U-Net \citep{ronneberger2015unet}        & 0.512 & 0.398 & 0.710 \\
U-Net++ \citep{zhou2018unetpp}           & 0.483 & 0.401 & 0.707 \\
ACSNet \citep{zhang2020acsnet}           & 0.716 & 0.578 & 0.863 \\
MSEG \citep{huang2021hardnetmseg}        & 0.735 & 0.700 & 0.874 \\
Polyp-PVT \citep{dong2023polyppvt}       & 0.808 & 0.787 & 0.900 \\
SAM-Mamba \citep{dutta2025sammamba}      & 0.853 & 0.848 & 0.920 \\
\midrule
This work: ResU-Net                      & 0.724 & 0.618 & 0.794 \\
This work: plain U-Net                   & 0.738 & 0.632 & 0.841 \\
This work: DINOv2 + LoRA $r{=}32$        & 0.769 & 0.698 & 0.867 \\
\bottomrule
\end{tabular*}
\end{table}

The baselines exceed the published U-Net by 0.226 on CVC-ColonDB and 0.234 on ETIS for the plain variant, and both are ahead of ACSNet on CVC-ColonDB. Since the plain U-Net and the published one are close in architecture, almost all of that difference is attributable to the training recipe rather than the model, which is the effect \citet{isensee2021nnunet} identify: the recipe yields a baseline more than twenty Dice points stronger than the conventional one.

Specialised polyp architectures published since 2023 exceed every model here, by 0.084 on CVC-ColonDB in the case of SAM-Mamba. This is expected: those architectures pair large pretrained encoders with multi-scale attention decoders developed for polyp segmentation, whereas the adapted model carries a 385-parameter linear decoder, chosen so that the adaptation axis measures how much the encoder must change rather than how much a decoder can compensate (Section~\ref{sec:adapt}).

The comparison made here is between two training paradigms under one controlled protocol, not a claim to the state of the art. What Table~\ref{tab:published} establishes is the narrower point the comparison of Section~\ref{sec:main} requires: the task-specific comparator is competitive with the published U-Net family it represents, so the deficit is not an artefact of a weak baseline.

\subsection{Failure modes, lesion size and boundary quality}\label{sec:failure}

Averages conceal how a model fails. Table~\ref{tab:empty} counts the images on which a model predicts no lesion at all, although every reference mask in every test set contains one. The plain U-Net never does so, and the residual variant on four polyp images. DINOv2 at the selected configuration misses 13 polyp images entirely, ten of them in CVC-ColonDB, and two breast images; elsewhere on the capacity axis the count is higher, between 25 and 40 of the 380 CVC-ColonDB images for every configuration other than LoRA at rank 32, and it rises steeply as training data shrink, to 58 CVC-ColonDB and 38 ETIS images at 10 per cent of the polyp data, against one image for the residual U-Net trained on the same subset. The missed polyps are small: the 94 CVC-ColonDB and ETIS images on which some configuration of the capacity axis predicts nothing have a median lesion area of 0.86 per cent of the image, 64 per cent of them fall in the smallest tercile of Table~\ref{tab:size}, and on them the plain U-Net reaches a mean DSC of 0.37. Skin, whose lesions occupy a large fraction of the image, has at most one empty prediction per configuration. Complete failure is not the commoner kind overall. Counting out-of-distribution images with DSC below 0.1, DINOv2 has 48 in polyp against 54 and 69 for the plain and residual U-Nets, and 31 in breast against 57 and 60. The adapted model therefore fails less often, but when it fails on a small lesion it tends to miss the lesion altogether, where a task-specific model produces a partial segmentation.

\begin{figure*}[tp]
\centering
\includegraphics[width=0.86\textwidth,height=0.74\textheight,keepaspectratio]{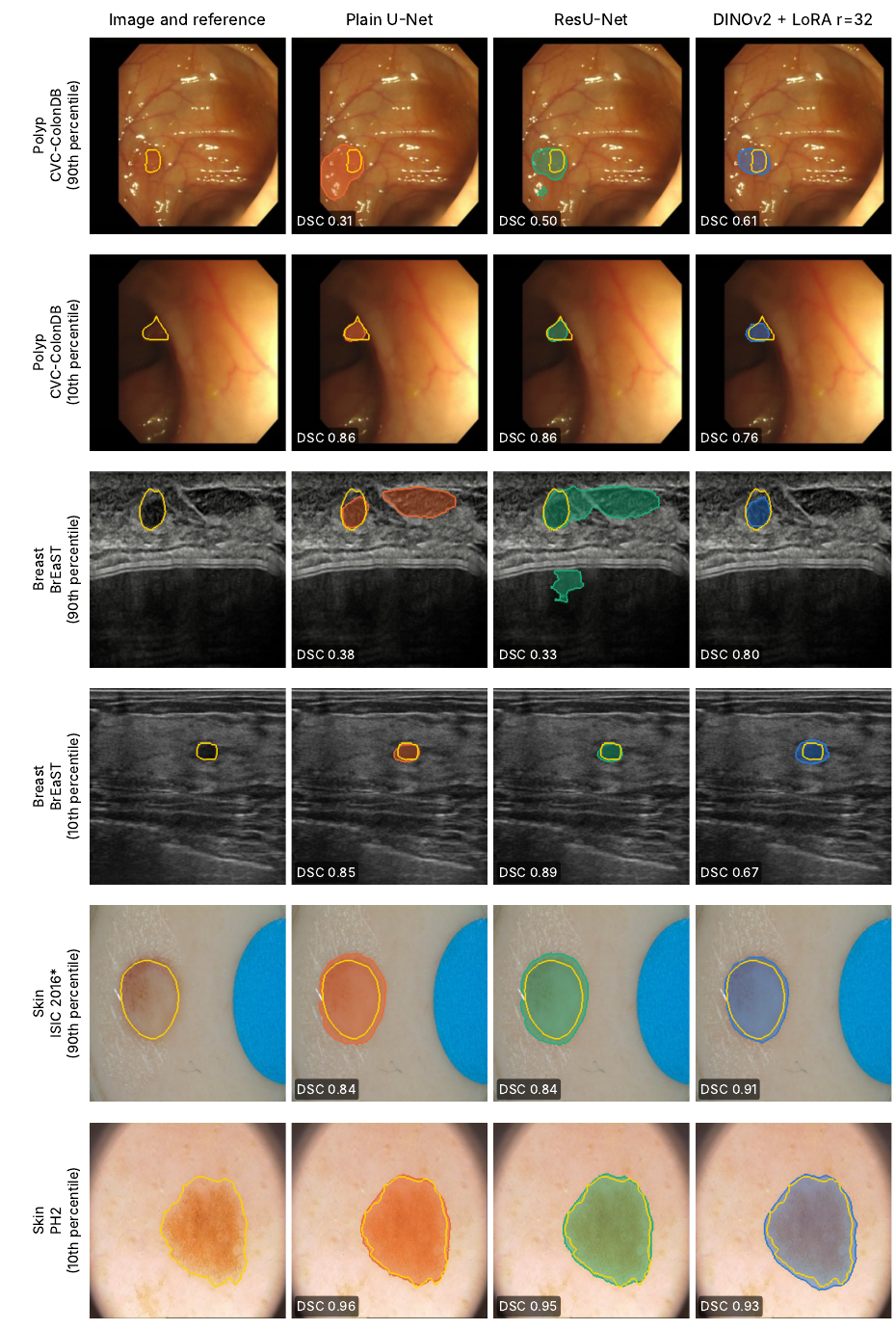}
\caption{Out-of-distribution cases selected by rule: in each domain, the images at the 90th and 10th percentiles of the per-image difference DSC(DINOv2 + LoRA $r{=}32$) $-$ DSC(plain U-Net), pooled over that domain's out-of-distribution sets. Yellow: reference contour; filled region and outline: model prediction, with its DSC. Images and masks are shown at $256{\times}256$; predictions are resampled from 224\,px (polyp, breast), 384\,px (skin U-Nets) and 392\,px (skin DINOv2, seed 42).}
\label{fig:qual}
\end{figure*}

Fig.~\ref{fig:qual} shows cases selected by a fixed rule rather than by eye: in each domain, the out-of-distribution images at the 90th and the 10th percentile of the per-image difference in DSC between DINOv2 and the plain U-Net. Where DINOv2 gains, the task-specific models also segment structures away from the lesion, the loss of precision quantified below; where it loses, all three models find the lesion and the difference lies in the boundary, the DINOv2 contour being coarser, most visibly at 224 pixels.

\begin{table*}[tp]
\centering
\caption{Images on which a model predicts no lesion at all, although every reference mask contains one. Task-specific models as in Table~\ref{tab:main}; capacity-axis configurations at 40 epochs in polyp and breast and 20 in skin, seed 42; data-size runs as in Section~\ref{sec:datasize}. --: configuration not run.}
\label{tab:empty}
\footnotesize
\setlength{\tabcolsep}{2pt}
\begin{adjustbox}{max width=\textwidth}
\begin{tabular}{@{}lccccccccccc@{}}
\toprule
 & \multicolumn{5}{c}{Polyp} & \multicolumn{3}{c}{Breast} & \multicolumn{3}{c}{Skin} \\
\cmidrule(lr){2-6}\cmidrule(lr){7-9}\cmidrule(l){10-12}
Configuration & Kvasir & ClinicDB & ColonDB & ETIS & CVC-T & BUSI-val & BrEaST & UCLM & 2016* & 2017* & PH2 \\
\midrule
Test images & 100 & 62 & 380 & 196 & 60 & 54 & 252 & 260 & 269 & 172 & 200 \\
\midrule
\multicolumn{12}{@{}l}{\textit{Task-specific models (Table~\ref{tab:main})}} \\
Plain U-Net & 0 & 0 & 0 & 0 & 0 & 0 & 0 & 0 & 0 & 0 & 0 \\
ResU-Net & 0 & 1 & 1 & 2 & 0 & 0 & 0 & 0 & 0 & 0 & 0 \\
\midrule
\multicolumn{12}{@{}l}{\textit{Capacity axis (Table~\ref{tab:capacity})}} \\
Frozen & 1 & 2 & 26 & 12 & 1 & 1 & 5 & 0 & -- & -- & -- \\
BitFit & 0 & 1 & 32 & 8 & 0 & 0 & 0 & 0 & 0 & 0 & 0 \\
LoRA $r{=}4$ & 1 & 1 & 35 & 7 & 0 & 1 & 0 & 1 & 0 & 0 & 0 \\
LoRA $r{=}8$ & 1 & 1 & 40 & 4 & 0 & 1 & 10 & 1 & 0 & 1 & 0 \\
LoRA $r{=}16$ & 1 & 1 & 30 & 10 & 0 & 0 & 2 & 0 & 0 & 1 & 0 \\
LoRA $r{=}32$ & 0 & 1 & 10 & 2 & 0 & 0 & 1 & 1 & 0 & 0 & 0 \\
Full FT & 1 & 1 & 25 & 9 & 0 & 0 & 6 & 3 & 0 & 0 & 0 \\
\midrule
\multicolumn{12}{@{}l}{\textit{Data-size sweep (Table~\ref{tab:datasize})}} \\
LoRA $r{=}32$, 10\% data & 2 & 3 & 58 & 38 & 2 & -- & -- & -- & -- & -- & -- \\
LoRA $r{=}32$, 25\% data & 0 & 2 & 25 & 11 & 0 & 0 & 0 & 3 & -- & -- & -- \\
LoRA $r{=}32$, 50\% data & 0 & 1 & 15 & 3 & 0 & 0 & 0 & 5 & -- & -- & -- \\
ResU-Net, 10\% data & 0 & 0 & 0 & 1 & 0 & -- & -- & -- & -- & -- & -- \\
ResU-Net, 25\% data & 0 & 0 & 5 & 0 & 0 & 0 & 0 & 0 & -- & -- & -- \\
ResU-Net, 50\% data & 0 & 0 & 1 & 2 & 0 & 0 & 0 & 0 & -- & -- & -- \\
\bottomrule
\end{tabular}
\end{adjustbox}
\end{table*}

Sensitivity and precision separate the remaining errors. Averaged over each domain's out-of-distribution sets, the DSC advantage of DINOv2 over the plain U-Net comes mainly from precision in polyp and breast, $+0.047$ and $+0.084$ against $+0.011$ and $+0.047$ in sensitivity, and mainly from sensitivity in skin, $+0.025$ against $+0.007$. Out of distribution, then, the task-specific models more often extend into tissue that is not lesion, rather than more often missing lesion tissue.

\begin{table}[tbp]
\centering
\caption{Out-of-distribution DSC by lesion size. Images are split into terciles of reference lesion area (per cent of the image) within each domain, pooled over its out-of-distribution sets. Plain: plain U-Net; ResU: ResU-Net; LoRA: DINOv2 + LoRA $r{=}32$. $\Delta$: LoRA minus the better task-specific model. Models and budgets as in Table~\ref{tab:main}.}
\label{tab:size}
\footnotesize
\setlength{\tabcolsep}{3pt}
\begin{tabular*}{\columnwidth}{@{\extracolsep{\fill}}llcrcccc@{}}
\toprule
Domain & Lesion & Area (\%) & $n$ & Plain & ResU & LoRA & $\Delta$ \\
\midrule
Polyp & Small & 0.1--2.0 & 212 & 0.542 & 0.531 & \textbf{0.560} & $+0.019$ \\
 & Medium & 2.0--5.0 & 212 & 0.819 & 0.804 & \textbf{0.870} & $+0.051$ \\
 & Large & 5.0--63.1 & 212 & 0.785 & 0.759 & \textbf{0.840} & $+0.055$ \\
\midrule
Breast & Small & 0.2--2.8 & 171 & 0.547 & 0.507 & \textbf{0.612} & $+0.065$ \\
 & Medium & 2.8--7.2 & 171 & 0.676 & 0.636 & \textbf{0.740} & $+0.064$ \\
 & Large & 7.2--46.2 & 170 & 0.707 & 0.723 & \textbf{0.775} & $+0.051$ \\
\midrule
Skin & Small & 0.9--13.6 & 214 & 0.880 & 0.875 & \textbf{0.893} & $+0.013$ \\
 & Medium & 13.6--32.0 & 213 & 0.899 & 0.887 & \textbf{0.915} & $+0.016$ \\
 & Large & 32.1--99.7 & 214 & 0.899 & 0.899 & \textbf{0.922} & $+0.023$ \\
\bottomrule
\end{tabular*}
\end{table}

Table~\ref{tab:size} stratifies out-of-distribution DSC by lesion area. DINOv2 leads in every stratum of every domain, but not evenly. In polyp its advantage over the better task-specific model is $+0.019$ for the smallest third of lesions against $+0.051$ and $+0.055$ for the larger two; in breast it is roughly constant, between $+0.051$ and $+0.065$; in skin it grows with lesion size from $+0.013$ to $+0.023$. Small lesions remain the hardest for every model, at a DSC between 0.51 and 0.61 in polyp and breast.

\begin{table*}[tp]
\centering
\caption{Boundary quality in skin at two resolutions: DINOv2 + LoRA $r{=}32$ (L) against ResU-Net (R). The 224\,px runs are the 100\% fraction of the data-size sweep, in which budgets were scaled (LoRA 20 epochs, ResU-Net 80); the native runs are those of Table~\ref{tab:main} (LoRA three-seed mean, 20 epochs; ResU-Net 100 epochs), so the comparison changes resolution together with these budgets rather than resolution alone. HD95 divided by the image diagonal. Bold marks the better model in each pair.}
\label{tab:res}
\footnotesize
\begin{tabular*}{\textwidth}{@{\extracolsep{\fill}}lcccccccc@{}}
\toprule
 & \multicolumn{2}{c}{DSC} & \multicolumn{2}{c}{NSD (2/224)} & \multicolumn{2}{c}{NSD (4/224)} & \multicolumn{2}{c}{HD95} \\
\cmidrule(lr){2-3}\cmidrule(lr){4-5}\cmidrule(lr){6-7}\cmidrule(l){8-9}
Test set & L & R & L & R & L & R & L & R \\
\midrule
\multicolumn{9}{@{}l}{\textit{224\,px for both models (data-size sweep, 100\%)}} \\
ISIC 2016* & \textbf{0.920} & 0.910 & 0.432 & \textbf{0.459} & \textbf{0.676} & 0.663 & \textbf{0.031} & 0.038 \\
ISIC 2017* & \textbf{0.859} & 0.856 & 0.402 & \textbf{0.415} & \textbf{0.581} & 0.576 & \textbf{0.053} & 0.054 \\
PH2 & \textbf{0.918} & 0.917 & 0.380 & \textbf{0.402} & 0.591 & \textbf{0.626} & \textbf{0.033} & 0.038 \\
\midrule
\multicolumn{9}{@{}l}{\textit{Native resolution, 392\,px (L) and 384\,px (R) (Table~\ref{tab:main})}} \\
ISIC 2016* & \textbf{0.924} & 0.908 & \textbf{0.431} & 0.415 & \textbf{0.698} & 0.644 & \textbf{0.029} & 0.042 \\
ISIC 2017* & \textbf{0.872} & 0.836 & \textbf{0.399} & 0.384 & \textbf{0.599} & 0.553 & \textbf{0.047} & 0.068 \\
PH2 & \textbf{0.925} & 0.903 & \textbf{0.386} & 0.308 & \textbf{0.636} & 0.531 & \textbf{0.031} & 0.048 \\
\bottomrule
\end{tabular*}
\end{table*}

The boundary results suggest an explanation in terms of resolution. DINOv2 represents a 224-pixel image as a $16{\times}16$ grid of patch embeddings, which the linear decoder upsamples bilinearly (Section~\ref{sec:adapt}), so its boundary can be no finer than that grid allows, whereas the U-Nets predict at full resolution. Skin, at 392 pixels, gives a $28{\times}28$ grid, and there DINOv2 leads in NSD on all three sets. Skin also allows a comparison within one domain, because its data-size sweep was run at 224 pixels (Table~\ref{tab:res}). At 224 pixels the residual U-Net leads in NSD at the 2/224 tolerance on all three skin sets while DINOv2 still leads in DSC and HD95; at the native resolutions DINOv2 leads in all of them. Within a domain, the direction of the fine-boundary comparison therefore changes with resolution, which is consistent with the patch grid limiting boundary precision. It is not an isolated test: the two pairs also differ in training budget, and at the lower data fractions of the same sweep the NSD comparison is mixed. The empty predictions on small polyps fit the same account, a lesion of one per cent of a 224-pixel image covering between two and three of its 256 patches, but this was not tested directly.

\subsection{More data closes the in-domain gap but not the out-of-distribution one}\label{sec:datasize}

Two explanations were left open for the variation in deficit magnitude: the domains differ in training-set size, and in the severity of the shift their test sets represent. This section addresses the first. Each domain was retrained at 25 and 50 per cent of its training set, and polyp additionally at 10 per cent, with source, architecture, evaluation and augmentation fixed, and budgets scaled inversely with subset size, which isolates data volume from modality and task difficulty as the cross-domain comparison cannot.

The cross-domain figures invite a straightforward reading: ordered by training-set size and by out-of-distribution deficit the domains fall in inverse order, so the deficit appears to shrink as data grows. That reading is incorrect, since the domains differ in modality, lesion appearance, annotation protocol and test-set provenance, any of which could produce the same ordering.

\begin{table*}[tbp]
\centering
\begin{minipage}[t]{\pairwidth}
\centering
\caption{Data-size sweep. Gaps are ResU-Net minus DINOv2 + LoRA $r{=}32$ on the identical subset, single-seed; positive favours training from scratch. In-domain and OOD gaps are averaged over each domain's in-domain and out-of-distribution sets respectively. Skin was run at $224{\times}224$ at every fraction (Section~\ref{sec:training}), so its gaps are not comparable with Table~\ref{tab:main} and are compared only within the sweep.}
\label{tab:datasize}
\footnotesize
\begin{tabular*}{\linewidth}{@{\extracolsep{\fill}}llrcc@{}}
\toprule
Domain & Fraction & Images & In-domain gap & OOD gap \\
\midrule
Polyp  & 10\%  &   130 & $-0.101$ & $-0.191$ \\
       & 25\%  &   326 & $-0.050$ & $-0.135$ \\
       & 50\%  &   652 & $-0.042$ & $-0.132$ \\
       & 100\% & 1,305 & $-0.006$ & $-0.066$ \\
\midrule
Breast & 25\%  &   119 & $-0.004$ & $-0.104$ \\
       & 50\%  &   239 & $+0.003$ & $-0.070$ \\
       & 100\% &   479 & $+0.021$ & $-0.087$ \\
\midrule
Skin   & 25\%  &   518 & -- & $-0.013$ \\
       & 50\%  & 1,037 & -- & $+0.001$ \\
       & 100\% & 2,075 & -- & $-0.005$ \\
\bottomrule
\end{tabular*}
\end{minipage}\pairsep
\begin{minipage}[t]{\pairwidth}
\centering
\caption{Configuration search, polyp at 326 images, single seed, from runs separate from those in Table~\ref{tab:datasize}. DINOv2 + LoRA $r{=}32$ on the same subset reaches 0.8435 in domain and 0.7439 out of distribution. Bold marks the best value in each column.}
\label{tab:configsearch}
\footnotesize
\setlength{\tabcolsep}{3pt}
\begin{tabular*}{\linewidth}{@{\extracolsep{\fill}}lrccc@{}}
\toprule
Configuration & Parameters & In-domain & OOD & OOD gap \\
\midrule
Baseline                & 8,114,692 & 0.7954 & 0.5849 & $-0.159$ \\
Reduced capacity        & 2,031,364 & 0.7789 & 0.5648 & $-0.179$ \\
Decoder dropout         & 8,114,692 & 0.7824 & 0.5780 & $-0.166$ \\
Heavier weight decay    & 8,114,692 & 0.7954 & 0.6019 & $-0.142$ \\
Aggressive augmentation & 8,114,692 & \textbf{0.8034} & \textbf{0.6414} & $\mathbf{-0.103}$ \\
\bottomrule
\end{tabular*}
\end{minipage}
\end{table*}

\begin{figure}[t]
\centering
\includegraphics[width=\linewidth]{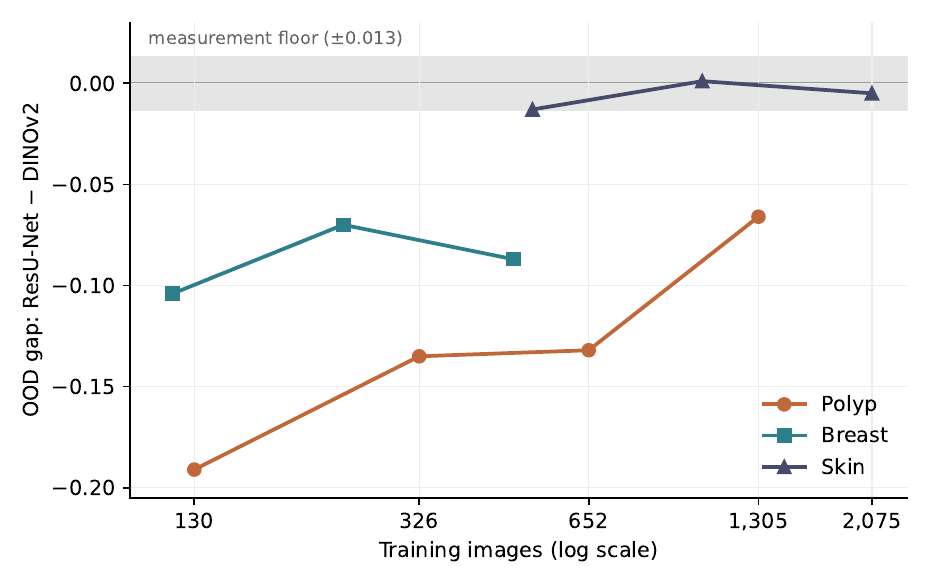}
\caption{Out-of-distribution gap between ResU-Net and DINOv2 + LoRA $r{=}32$ against training-set size, by domain. Negative values indicate the task-specific model is behind; the shaded band marks the 0.013 measurement floor.}
\label{fig:datasize}
\end{figure}

Comparing points of equal size across domains, as Fig.~\ref{fig:datasize} shows, contradicts the confounded reading directly. At 518 images skin shows $-0.013$, at the boundary of the floor, while at 652 images, twenty-six per cent more data, polyp shows $-0.132$, ten times larger. The relationship does not merely weaken across domains; it reverses. Within domains it does hold, falling monotonically in polyp and less regularly in breast, and remaining indistinguishable from zero in skin at every fraction. The correct statement is conditional: within a domain more data reduces the deficit; across domains data volume does not predict it.

The in-domain column tells a different story, and the contrast is the substantive finding of this section. Three of the seven configurations with an in-domain measurement fall within the floor, polyp at its full training set, and breast at 119 and 239 images, and at 479 images the task-specific model in breast is ahead by $+0.021$, a margin that clears it on the 54-image validation split. Only polyp requires its full set to reach parity, and even at a quarter of its data the in-domain gap is $-0.050$ against $-0.135$ out of distribution on the same models. The plain U-Net reproduces this, its gaps moving from $-0.034$ and $-0.121$ at 326 images to $+0.002$ and $-0.041$ at 1,305, so the behaviour is not specific to one model; the 326-image run stopped at epoch 367 of its 1,200-epoch budget, so its deficit is not a training artefact. Stated directly: a task-specific model reaches parity on data from the training distribution using, in breast, a fraction of the available data, and does not reach it on data from a different source using all of it.

Data volume is one candidate explanation for the out-of-distribution deficit; regularisation is another, since a model trained on a small dataset might be overfitting in a way a better-configured version would not. A configuration search on polyp at 326 images, in runs separate from those in Table~\ref{tab:datasize}, tested this (Table~\ref{tab:configsearch}). Reducing capacity to a quarter of the parameters hurt on both regimes, which argues against overfitting as the explanation; dropout hurt; heavier weight decay helped modestly, by $+0.017$ out of distribution; and aggressive augmentation was the clear winner at $+0.057$ and was adopted for all subsequent task-specific runs. None of the five closed the gap: the best remains 0.103 behind the adapted model trained on the same 326 images.

The trend holds in the distance metric and changes character at the boundary (Table~\ref{tab:sizefull}). In polyp the out-of-distribution HD95 favours adaptation at every fraction, while the out-of-distribution NSD gap shrinks from $-0.175$ at 10 per cent to $-0.011$ at the full set, and in domain NSD favours the residual U-Net from 50 per cent upward, by up to $+0.079$. In breast the out-of-distribution NSD gap stays within 0.02 of zero at every fraction, so there the out-of-distribution advantage of adaptation lies in locating the lesion rather than in tracing its boundary.

\begin{table*}[tbp]
\centering
\caption{Data-size sweep in three metrics: ResU-Net minus DINOv2 + LoRA $r{=}32$ on the identical subset, averaged over each domain's in-domain and out-of-distribution sets (DSC as in Table~\ref{tab:datasize}). Positive DSC and NSD gaps and negative HD95 gaps favour training from scratch. Skin at 224 px at every fraction; the 100\% rows of polyp and breast are the models of Table~\ref{tab:main}.}
\label{tab:sizefull}
\footnotesize
\begin{tabular*}{\textwidth}{@{\extracolsep{\fill}}llcccccc@{}}
\toprule
 & & \multicolumn{3}{c}{In-domain gap} & \multicolumn{3}{c}{Out-of-distribution gap} \\
\cmidrule(lr){3-5}\cmidrule(lr){6-8}
Domain & Fraction & DSC & HD95 & NSD & DSC & HD95 & NSD \\
\midrule
Polyp & 10\% & $-0.101$ & $+0.036$ & $-0.036$ & $-0.191$ & $+0.054$ & $-0.175$ \\
 & 25\% & $-0.050$ & $+0.021$ & $+0.008$ & $-0.135$ & $+0.071$ & $-0.126$ \\
 & 50\% & $-0.042$ & $+0.014$ & $+0.049$ & $-0.132$ & $+0.066$ & $-0.096$ \\
 & 100\% & $-0.006$ & $+0.005$ & $+0.079$ & $-0.066$ & $+0.042$ & $-0.011$ \\
\midrule
Breast & 25\% & $-0.004$ & $+0.021$ & $+0.125$ & $-0.104$ & $+0.094$ & $-0.007$ \\
 & 50\% & $+0.003$ & $+0.015$ & $+0.086$ & $-0.070$ & $+0.065$ & $-0.003$ \\
 & 100\% & $+0.021$ & $+0.003$ & $+0.133$ & $-0.087$ & $+0.084$ & $+0.016$ \\
\midrule
Skin & 25\% & -- & -- & -- & $-0.013$ & $+0.007$ & $-0.009$ \\
 & 50\% & -- & -- & -- & $+0.001$ & $+0.000$ & $+0.032$ \\
 & 100\% & -- & -- & -- & $-0.005$ & $+0.004$ & $+0.021$ \\
\bottomrule
\end{tabular*}
\end{table*}

Two conclusions follow, pointing in different directions. Adapting a foundation model confers no advantage in fitting the training distribution once enough data is available, which polyp reaches only at its full set; below that point adaptation does help in domain, by 0.101 at 130 polyp images, 0.050 at 326 and 0.042 at 652, while in breast the scores are already close at 119 images, the smallest subset reported, though that comparison rests on the validation split rather than a held-out test set. And the variation in deficit magnitude across domains is not explained by training-set size, so whatever distinguishes the domains is a property of the domain rather than of how much data it provides. Section~\ref{sec:predictors} reports three attempts to identify it, none of which succeeded.

\subsection{Adaptation capacity requirements differ by domain}\label{sec:capacity}

Given that a foundation model is warranted where cross-site deployment is intended, the remaining question is how much adaptation it requires. This section reports four experiments.

\subsubsection{Parameter-efficient adaptation matches full fine-tuning}

LoRA at rank 32 was compared against full fine-tuning in polyp across three seeds at 20 epochs, with Wilcoxon signed-rank tests on seed-averaged per-image vectors (Table~\ref{tab:lorafull}). LoRA minus full fine-tuning was $+0.0058$ on Kvasir ($p=0.0003$), $+0.0101$ on CVC-ClinicDB ($p=0.0194$), $+0.0022$ on CVC-ColonDB ($p=0.0001$), $+0.0130$ on ETIS ($p=0.1191$) and $+0.0006$ on CVC-T ($p=0.3931$); for the last two the test records a failure to reject, not equivalence. LoRA is equal or ahead on all five, significantly on three, and full fine-tuning wins nowhere despite updating thirty-seven times as many parameters. The magnitudes are small and fall at or below the floor governing single-run comparisons, but that floor does not apply to a paired test over images whose scores are averaged across three seeds. Skin gives comparable differences, significant on two of its three sets at the same rank. This is consistent with the wider literature and is not novel; it is reported because the rest of the section depends on it. In the other metrics the comparison is not one-sided (Table~\ref{tab:seedsfull}). LoRA is also ahead on all five sets in IoU and NSD, significantly on three in each, and has significantly higher sensitivity on four, but its precision is significantly lower on all three out-of-distribution sets, by up to 0.049 on CVC-ColonDB: at equal or better DSC it covers more of each polyp and more of its surroundings than full fine-tuning does. LoRA at rank 32 is therefore equal or ahead in DSC, IoU and NSD, not in every metric.

\begin{table*}[tbp]
\centering
\caption{LoRA $r{=}32$ against full fine-tuning, polyp, 20 epochs, three seeds. Values are three-seed means (seeds 42, 123 and 2024); the seed-42 LoRA $r{=}32$ run alone appears in Table~\ref{tab:trunc}. $\Delta$ is LoRA minus full fine-tuning; the verdict applies the criterion of Section~\ref{sec:eval}, not the measurement floor. ``No difference'' records a failure to reject, not equivalence.}
\label{tab:lorafull}
\footnotesize
\begin{tabular*}{\textwidth}{@{\extracolsep{\fill}}lccccl@{}}
\toprule
Test set & LoRA $r{=}32$ & Full FT & $\Delta$ & $p$ & Verdict \\
\midrule
Kvasir (in-domain)       & 0.8825 & 0.8767 & $+0.0058$ & 0.0003 & LoRA \\
CVC-ClinicDB (in-domain) & 0.8485 & 0.8384 & $+0.0101$ & 0.0194 & LoRA \\
CVC-ColonDB              & 0.7401 & 0.7380 & $+0.0022$ & 0.0001 & LoRA \\
ETIS                     & 0.6986 & 0.6856 & $+0.0130$ & 0.1191 & No difference \\
CVC-T                    & 0.8671 & 0.8665 & $+0.0006$ & 0.3931 & No difference \\
\bottomrule
\end{tabular*}
\end{table*}

\begin{table*}[tbp]
\centering
\caption{LoRA against full fine-tuning in all metrics, three seeds (42, 123, 2024), 20 epochs; polyp as in Table~\ref{tab:lorafull}, skin as in Table~\ref{tab:rank}. Each entry is the mean over images of the seed-averaged per-image difference, LoRA minus full fine-tuning; for HD95 (divided by the image diagonal) a negative value favours LoRA, for all other metrics a positive one. An asterisk marks $p < 0.05$ in a Wilcoxon signed-rank test on the seed-averaged per-image vectors (Section~\ref{sec:eval}).}
\label{tab:seedsfull}
\footnotesize
\begin{tabular*}{\textwidth}{@{\extracolsep{\fill}}lcccccc@{}}
\toprule
Test set & DSC & IoU & Sens & Prec & HD95 & NSD \\
\midrule
\multicolumn{7}{@{}l}{\textit{Polyp, LoRA $r{=}32$ $-$ full FT}} \\
Kvasir & $+0.0058^{*}$ & $+0.0103^{*}$ & $+0.0041$ & $+0.0146^{*}$ & $-0.0059$ & $+0.0451^{*}$ \\
CVC-ClinicDB & $+0.0101^{*}$ & $+0.0151^{*}$ & $+0.0166^{*}$ & $-0.0005$ & $-0.0018$ & $+0.0399^{*}$ \\
CVC-ColonDB & $+0.0022^{*}$ & $+0.0121^{*}$ & $+0.0180^{*}$ & $-0.0486^{*}$ & $+0.0284$ & $+0.0264^{*}$ \\
ETIS & $+0.0130$ & $+0.0133$ & $+0.0416^{*}$ & $-0.0147^{*}$ & $-0.0028$ & $+0.0130$ \\
CVC-T & $+0.0007$ & $+0.0040$ & $+0.0391^{*}$ & $-0.0319^{*}$ & $+0.0022$ & $+0.0049$ \\
\midrule
\multicolumn{7}{@{}l}{\textit{Skin, LoRA $r{=}8$ $-$ full FT}} \\
ISIC 2016* & $-0.0035^{*}$ & $-0.0061^{*}$ & $-0.0088^{*}$ & $+0.0021$ & $+0.0022^{*}$ & $-0.0187^{*}$ \\
ISIC 2017* & $-0.0168^{*}$ & $-0.0211^{*}$ & $-0.0081$ & $-0.0187^{*}$ & $+0.0097^{*}$ & $-0.0249^{*}$ \\
PH2 & $+0.0013^{*}$ & $+0.0030^{*}$ & $-0.0073^{*}$ & $+0.0099^{*}$ & $+0.0012$ & $+0.0268^{*}$ \\
\midrule
\multicolumn{7}{@{}l}{\textit{Skin, LoRA $r{=}32$ $-$ full FT}} \\
ISIC 2016* & $+0.0027^{*}$ & $+0.0048^{*}$ & $+0.0023$ & $+0.0026$ & $-0.0017^{*}$ & $+0.0172^{*}$ \\
ISIC 2017* & $-0.0018$ & $-0.0005$ & $+0.0017$ & $-0.0019$ & $+0.0019$ & $+0.0083$ \\
PH2 & $+0.0032^{*}$ & $+0.0062^{*}$ & $+0.0001$ & $+0.0063^{*}$ & $-0.0019^{*}$ & $+0.0326^{*}$ \\
\bottomrule
\end{tabular*}
\end{table*}

\subsubsection{The rank requirement differs by domain}

Swept over ranks 4, 8, 16 and 32 under the 20-epoch protocol, the domains give three profiles. Skin gains nothing from rank beyond 4, no set moving beyond the floor and the largest change to rank 32 being $+0.005$ on ISIC 2017*; breast improves to rank 16 then plateaus; polyp continues improving at 32. Expressed as a requirement, skin needs approximately rank 4, breast 16 and polyp at least 32, a factor of eight, so no single default serves all three. One feature of the training logs qualifies this: at 20 epochs the low ranks were still improving while the high ranks had converged, so part of what is measured as capacity may be a convergence rate.

\subsubsection{The effect of the epoch budget}\label{sec:budget}

The qualification at the end of the previous section is testable, and this experiment tests it. If the low ranks were still improving at 20 epochs while the high ranks had converged, then extending the budget should shrink the measured rank effect where it reflects convergence and leave it where it reflects capacity. Polyp and breast were therefore rerun at 40 epochs with all else unchanged, and the rank effect was measured as each domain's span at 20 epochs, rank 16 minus rank 4 in breast and rank 32 minus rank 4 in polyp (Table~\ref{tab:budgetctl}).

\begin{table}[tbp]
\centering
\caption{Effect of doubling the epoch budget on the measured rank effect. Breast rows use $r{=}16 - r{=}4$ and the polyp row $r{=}32 - r{=}4$, in each case that domain's span at 20 epochs. Single seed.}
\label{tab:budgetctl}
\footnotesize
\setlength{\tabcolsep}{3pt}
\begin{tabular*}{\columnwidth}{@{\extracolsep{\fill}}lllcc@{}}
\toprule
Domain & Test set & Regime & 20 epochs & 40 epochs \\
\midrule
Breast & BrEaST      & OOD       & $+0.030$ & $+0.001$ \\
Breast & BUS-UCLM    & OOD       & $+0.023$ & $+0.014$ \\
Breast & BUSI-val    & in-domain & $+0.040$ & $+0.038$ \\
\midrule
Polyp  & CVC-ColonDB & OOD       & $+0.046$ & $+0.055$ \\
\bottomrule
\end{tabular*}
\end{table}

In breast the out-of-distribution rank effect largely dissolves. It vanishes on BrEaST, from $+0.030$ to $+0.001$, where rank 4 comes within 0.001 of rank 16 given twice the budget, and halves on BUS-UCLM, from $+0.023$ to $+0.014$, which sits at the edge of the floor. What appeared at 20 epochs as a requirement of rank 16 was therefore in substantial part a difference in convergence speed. In polyp it survives and grows slightly, from $+0.046$ to $+0.055$ on CVC-ColonDB. Convergence was verified rather than assumed, the rank-16 and full fine-tuning runs early-stopping at epochs 31 and 35 while the rest moved by no more than 0.001 over their final six epochs.

The in-domain measurement behaves differently within the same domain: doubling the budget leaves it almost unchanged, from $+0.040$ to $+0.038$, while the out-of-distribution effect largely dissolves, so the two were measuring different things at 20 epochs, in-domain reflecting capacity and out-of-distribution largely reflecting optimisation progress. The difference between reporting a requirement of 16 and one of 4 in breast rests entirely on whether the experiment ran for 20 or 40 epochs. All subsequent capacity analyses in polyp and breast are therefore at 40 epochs, skin remaining at 20 to match its rank sweep, and the domain-dependence survives in weakened form.

\subsubsection{The requirement is not specific to LoRA}

\begin{figure*}[t]
\centering
\includegraphics[width=\linewidth]{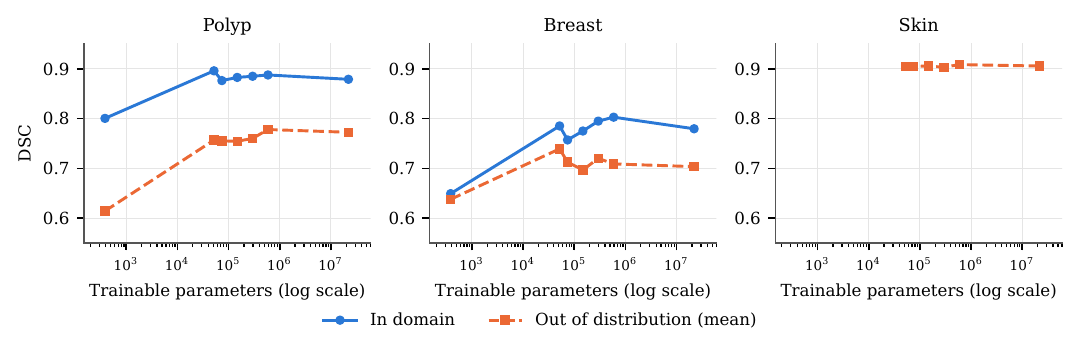}
\caption{Capacity axis by domain: DSC against trainable parameter count (logarithmic scale), with the values of Table~\ref{tab:capacity}. Polyp and breast at 40 epochs, skin at 20, all seed 42; from left to right the points are frozen (polyp and breast only), BitFit, LoRA at ranks 4, 8, 16 and 32, and full fine-tuning. Solid lines: in-domain DSC (Kvasir, BUSI-val); dashed lines: mean over the domain's out-of-distribution sets. The three panels share the vertical axis.}
\label{fig:capacity}
\end{figure*}

Rank belongs to LoRA's mechanism, and it remains possible that what varies is how well low-rank factorisation suits each domain's required update. BitFit tests this, having no rank parameter, intervening at bias terms throughout the encoder, and using 51,841 parameters, below LoRA at rank 4 (Table~\ref{tab:capacity}, Fig.~\ref{fig:capacity}).

\begin{table*}[p]
\centering
\caption{Capacity axis by test set in all metrics. Polyp and breast at 40 epochs, skin at 20 to match its rank sweep, all seed 42. The skin frozen configuration is omitted: its checkpoint was trained under an earlier protocol and is not comparable. DSC values are those of the original evaluation, except on CVC-ClinicDB, which was not reported there and is recomputed. HD95 and ASSD are divided by the image diagonal (lower is better); NSD$_{2}$ and NSD$_{4}$ at 2/224 and 4/224 of the image width. BitFit's BUSI-val value is its own selection criterion, not an independent result. (in) marks in-domain sets. Bold marks the best configuration for each test set and metric.}
\label{tab:capacity}
\scriptsize
\renewcommand{\arraystretch}{0.94}
\setlength{\tabcolsep}{2pt}
\providecommand{\CapTabMax}{\textheight}
\begin{adjustbox}{max width=\textwidth, max totalheight=\CapTabMax}
\begin{tabular}{@{}llrcccccccc@{}}
\toprule
Test set & Configuration & Trainable & DSC & IoU & Sens & Prec & HD95 & ASSD & NSD$_{2}$ & NSD$_{4}$ \\
\midrule
\multicolumn{11}{@{}l}{\textit{Polyp}} \\
Kvasir (in) & Frozen & 385 & 0.8005 & 0.693 & 0.826 & 0.822 & 0.118 & 0.037 & 0.345 & 0.546 \\
 & BitFit & 51,841 & \textbf{0.8960} & \textbf{0.829} & \textbf{0.888} & \textbf{0.926} & \textbf{0.056} & \textbf{0.015} & 0.628 & 0.820 \\
 & LoRA $r{=}4$ & 74,113 & 0.8764 & 0.805 & 0.878 & 0.899 & 0.070 & 0.025 & 0.587 & 0.797 \\
 & LoRA $r{=}8$ & 147,841 & 0.8828 & 0.816 & 0.874 & 0.917 & 0.069 & 0.025 & 0.618 & 0.813 \\
 & LoRA $r{=}16$ & 295,297 & 0.8851 & 0.819 & 0.872 & 0.923 & 0.065 & 0.024 & 0.626 & 0.816 \\
 & LoRA $r{=}32$ & 590,209 & 0.8877 & 0.823 & 0.883 & 0.918 & 0.058 & 0.016 & \textbf{0.649} & \textbf{0.821} \\
 & Full FT & 22,056,961 & 0.8790 & 0.814 & 0.866 & 0.921 & 0.066 & 0.025 & 0.616 & 0.811 \\
\addlinespace[2pt]
CVC-ClinicDB (in) & Frozen & 385 & 0.6962 & 0.585 & 0.740 & 0.753 & 0.148 & 0.065 & 0.331 & 0.531 \\
 & BitFit & 51,841 & 0.8306 & 0.754 & 0.858 & 0.843 & 0.059 & 0.029 & 0.578 & 0.795 \\
 & LoRA $r{=}4$ & 74,113 & 0.8218 & 0.739 & 0.872 & 0.819 & 0.062 & 0.031 & 0.531 & 0.753 \\
 & LoRA $r{=}8$ & 147,841 & 0.8490 & 0.773 & 0.857 & 0.881 & 0.054 & 0.027 & 0.603 & 0.819 \\
 & LoRA $r{=}16$ & 295,297 & 0.8527 & 0.777 & 0.850 & \textbf{0.889} & 0.054 & 0.027 & 0.629 & 0.821 \\
 & LoRA $r{=}32$ & 590,209 & \textbf{0.8684} & \textbf{0.798} & \textbf{0.892} & 0.871 & \textbf{0.046} & \textbf{0.026} & \textbf{0.653} & \textbf{0.838} \\
 & Full FT & 22,056,961 & 0.8566 & 0.780 & 0.855 & 0.887 & 0.049 & 0.026 & 0.629 & 0.821 \\
\addlinespace[2pt]
CVC-ColonDB & Frozen & 385 & 0.5875 & 0.476 & 0.601 & 0.685 & 0.196 & 0.109 & 0.250 & 0.424 \\
 & BitFit & 51,841 & 0.7204 & 0.635 & 0.740 & 0.752 & 0.139 & 0.105 & 0.462 & 0.662 \\
 & LoRA $r{=}4$ & 74,113 & 0.7139 & 0.627 & 0.753 & 0.738 & 0.147 & 0.112 & 0.427 & 0.645 \\
 & LoRA $r{=}8$ & 147,841 & 0.7132 & 0.632 & 0.740 & 0.751 & 0.157 & 0.123 & 0.458 & 0.665 \\
 & LoRA $r{=}16$ & 295,297 & 0.7346 & 0.650 & 0.768 & 0.760 & 0.133 & 0.098 & 0.482 & 0.689 \\
 & LoRA $r{=}32$ & 590,209 & \textbf{0.7694} & \textbf{0.683} & \textbf{0.816} & 0.780 & \textbf{0.086} & \textbf{0.049} & \textbf{0.515} & \textbf{0.719} \\
 & Full FT & 22,056,961 & 0.7396 & 0.651 & 0.748 & \textbf{0.790} & 0.118 & 0.084 & 0.486 & 0.690 \\
\addlinespace[2pt]
ETIS & Frozen & 385 & 0.4947 & 0.394 & 0.585 & 0.496 & 0.290 & 0.133 & 0.215 & 0.369 \\
 & BitFit & 51,841 & 0.6867 & 0.596 & 0.774 & 0.661 & 0.127 & 0.071 & 0.467 & 0.660 \\
 & LoRA $r{=}4$ & 74,113 & 0.6843 & 0.588 & 0.762 & 0.673 & 0.121 & 0.064 & 0.411 & 0.638 \\
 & LoRA $r{=}8$ & 147,841 & 0.6984 & 0.602 & 0.760 & 0.690 & 0.103 & 0.049 & 0.450 & 0.658 \\
 & LoRA $r{=}16$ & 295,297 & 0.6711 & 0.579 & 0.746 & 0.659 & 0.137 & 0.082 & 0.432 & 0.631 \\
 & LoRA $r{=}32$ & 590,209 & 0.6984 & 0.609 & \textbf{0.805} & 0.654 & \textbf{0.099} & \textbf{0.041} & \textbf{0.478} & 0.669 \\
 & Full FT & 22,056,961 & \textbf{0.7041} & \textbf{0.610} & 0.756 & \textbf{0.702} & 0.117 & 0.070 & 0.467 & \textbf{0.676} \\
\addlinespace[2pt]
CVC-T & Frozen & 385 & 0.7612 & 0.644 & 0.798 & 0.791 & 0.116 & 0.041 & 0.384 & 0.629 \\
 & BitFit & 51,841 & 0.8634 & 0.788 & 0.924 & 0.830 & 0.029 & 0.009 & 0.672 & 0.876 \\
 & LoRA $r{=}4$ & 74,113 & 0.8675 & 0.784 & \textbf{0.954} & 0.814 & \textbf{0.023} & 0.009 & 0.599 & 0.856 \\
 & LoRA $r{=}8$ & 147,841 & 0.8520 & 0.767 & 0.945 & 0.807 & 0.029 & 0.010 & 0.579 & 0.837 \\
 & LoRA $r{=}16$ & 295,297 & \textbf{0.8749} & \textbf{0.801} & 0.927 & 0.854 & 0.027 & 0.009 & 0.668 & \textbf{0.885} \\
 & LoRA $r{=}32$ & 590,209 & 0.8665 & 0.792 & 0.951 & 0.827 & 0.032 & 0.010 & 0.672 & 0.871 \\
 & Full FT & 22,056,961 & 0.8739 & 0.796 & 0.899 & \textbf{0.881} & 0.025 & \textbf{0.008} & \textbf{0.680} & 0.877 \\
\midrule
\multicolumn{11}{@{}l}{\textit{Breast}} \\
BUSI-val (in) & Frozen & 385 & 0.6493 & 0.524 & 0.731 & 0.652 & 0.161 & 0.062 & 0.230 & 0.399 \\
 & BitFit & 51,841 & 0.7853 & 0.688 & 0.776 & \textbf{0.833} & \textbf{0.067} & \textbf{0.025} & \textbf{0.447} & \textbf{0.675} \\
 & LoRA $r{=}4$ & 74,113 & 0.7572 & 0.654 & 0.796 & 0.782 & 0.098 & 0.043 & 0.368 & 0.601 \\
 & LoRA $r{=}8$ & 147,841 & 0.7751 & 0.675 & 0.812 & 0.785 & 0.089 & 0.042 & 0.408 & 0.619 \\
 & LoRA $r{=}16$ & 295,297 & 0.7950 & 0.694 & \textbf{0.846} & 0.779 & 0.072 & 0.028 & 0.411 & 0.646 \\
 & LoRA $r{=}32$ & 590,209 & \textbf{0.8029} & \textbf{0.709} & 0.844 & 0.788 & 0.072 & 0.028 & 0.438 & 0.661 \\
 & Full FT & 22,056,961 & 0.7796 & 0.681 & 0.813 & 0.769 & 0.077 & 0.028 & 0.406 & 0.638 \\
\addlinespace[2pt]
BrEaST & Frozen & 385 & 0.6099 & 0.485 & 0.658 & 0.671 & 0.142 & 0.058 & 0.222 & 0.386 \\
 & BitFit & 51,841 & \textbf{0.7337} & \textbf{0.620} & 0.833 & 0.723 & 0.076 & \textbf{0.024} & \textbf{0.362} & \textbf{0.570} \\
 & LoRA $r{=}4$ & 74,113 & 0.7165 & 0.598 & 0.802 & \textbf{0.724} & \textbf{0.073} & 0.024 & 0.309 & 0.516 \\
 & LoRA $r{=}8$ & 147,841 & 0.6957 & 0.582 & 0.770 & 0.710 & 0.106 & 0.063 & 0.304 & 0.511 \\
 & LoRA $r{=}16$ & 295,297 & 0.7179 & 0.605 & \textbf{0.843} & 0.693 & 0.079 & 0.032 & 0.328 & 0.530 \\
 & LoRA $r{=}32$ & 590,209 & 0.7133 & 0.604 & 0.813 & 0.711 & 0.075 & 0.029 & 0.338 & 0.541 \\
 & Full FT & 22,056,961 & 0.6941 & 0.579 & 0.786 & 0.702 & 0.095 & 0.049 & 0.307 & 0.501 \\
\addlinespace[2pt]
BUS-UCLM & Frozen & 385 & 0.6660 & 0.537 & 0.736 & 0.686 & 0.150 & 0.040 & 0.256 & 0.427 \\
 & BitFit & 51,841 & \textbf{0.7449} & \textbf{0.636} & 0.836 & \textbf{0.746} & \textbf{0.078} & \textbf{0.025} & \textbf{0.382} & \textbf{0.576} \\
 & LoRA $r{=}4$ & 74,113 & 0.7087 & 0.597 & 0.804 & 0.718 & 0.093 & 0.034 & 0.320 & 0.524 \\
 & LoRA $r{=}8$ & 147,841 & 0.6979 & 0.588 & 0.812 & 0.682 & 0.092 & 0.034 & 0.320 & 0.517 \\
 & LoRA $r{=}16$ & 295,297 & 0.7226 & 0.613 & \textbf{0.870} & 0.696 & 0.080 & 0.028 & 0.336 & 0.538 \\
 & LoRA $r{=}32$ & 590,209 & 0.7046 & 0.600 & 0.783 & 0.738 & 0.082 & 0.031 & 0.349 & 0.554 \\
 & Full FT & 22,056,961 & 0.7130 & 0.606 & 0.779 & 0.736 & 0.093 & 0.039 & 0.347 & 0.551 \\
\midrule
\multicolumn{11}{@{}l}{\textit{Skin}} \\
ISIC 2016* & BitFit & 51,841 & 0.9238 & 0.864 & \textbf{0.940} & 0.921 & 0.029 & 0.012 & 0.420 & 0.695 \\
 & LoRA $r{=}4$ & 74,113 & 0.9236 & 0.863 & 0.935 & 0.926 & 0.030 & 0.012 & 0.426 & 0.693 \\
 & LoRA $r{=}8$ & 147,841 & 0.9242 & 0.864 & 0.936 & 0.926 & 0.029 & \textbf{0.012} & 0.427 & 0.691 \\
 & LoRA $r{=}16$ & 295,297 & 0.9233 & 0.863 & 0.938 & 0.922 & 0.030 & 0.012 & 0.411 & 0.695 \\
 & LoRA $r{=}32$ & 590,209 & \textbf{0.9246} & \textbf{0.866} & 0.938 & 0.927 & \textbf{0.029} & 0.012 & \textbf{0.448} & \textbf{0.705} \\
 & Full FT & 22,056,961 & 0.9215 & 0.860 & 0.930 & \textbf{0.928} & 0.031 & 0.012 & 0.429 & 0.682 \\
\addlinespace[2pt]
ISIC 2017* & BitFit & 51,841 & 0.8684 & 0.786 & 0.838 & 0.944 & 0.049 & 0.021 & 0.396 & 0.583 \\
 & LoRA $r{=}4$ & 74,113 & 0.8710 & 0.792 & 0.842 & 0.944 & 0.047 & 0.020 & \textbf{0.406} & \textbf{0.612} \\
 & LoRA $r{=}8$ & 147,841 & 0.8635 & 0.785 & 0.840 & 0.936 & 0.052 & 0.025 & 0.394 & 0.587 \\
 & LoRA $r{=}16$ & 295,297 & 0.8629 & 0.783 & 0.830 & 0.944 & 0.052 & 0.025 & 0.380 & 0.590 \\
 & LoRA $r{=}32$ & 590,209 & \textbf{0.8763} & \textbf{0.798} & \textbf{0.847} & 0.948 & \textbf{0.044} & \textbf{0.019} & 0.404 & 0.603 \\
 & Full FT & 22,056,961 & 0.8717 & 0.791 & 0.836 & \textbf{0.951} & 0.045 & 0.020 & 0.389 & 0.589 \\
\addlinespace[2pt]
PH2 & BitFit & 51,841 & 0.9208 & 0.859 & 0.981 & 0.877 & 0.033 & 0.015 & 0.365 & 0.613 \\
 & LoRA $r{=}4$ & 74,113 & 0.9205 & 0.859 & 0.982 & 0.876 & 0.033 & 0.015 & 0.370 & 0.614 \\
 & LoRA $r{=}8$ & 147,841 & \textbf{0.9289} & \textbf{0.872} & 0.981 & \textbf{0.890} & \textbf{0.031} & \textbf{0.013} & \textbf{0.414} & \textbf{0.655} \\
 & LoRA $r{=}16$ & 295,297 & 0.9236 & 0.863 & \textbf{0.986} & 0.875 & 0.032 & 0.014 & 0.378 & 0.614 \\
 & LoRA $r{=}32$ & 590,209 & 0.9246 & 0.865 & 0.984 & 0.880 & 0.032 & 0.014 & 0.390 & 0.635 \\
 & Full FT & 22,056,961 & 0.9239 & 0.863 & 0.982 & 0.880 & 0.033 & 0.014 & 0.374 & 0.621 \\
\bottomrule
\end{tabular}
\end{adjustbox}
\end{table*}

The domain ordering reproduces under BitFit. Taking BitFit minus LoRA $r{=}32$ across the eight out-of-distribution sets gives $-0.001$, $-0.008$ and $-0.004$ in skin; $+0.020$ and $+0.040$ in breast; and $-0.003$, $-0.012$ and $-0.049$ in polyp. On seven of eight sets, 51,841 parameters match or exceed what 590,209 achieve, and the single deficit outside the floor is CVC-ColonDB. Polyp is thus the only domain where the eleven-fold larger budget buys anything, and it is the domain the rank sweep identified as requiring rank 32. Two methods differing in parameter count, in intervention site, and in whether they possess a rank parameter at all place the same domain above the others, which locates the requirement in the domain rather than in low-rank factorisation.

Two features of the axis are consistent where the full axis is available. The initial step carries most of the benefit: frozen to BitFit gains $+0.192$ on ETIS, $+0.133$ on CVC-ColonDB and $+0.102$ on CVC-T in polyp, and $+0.124$ and $+0.079$ in breast, and on every one the remaining step to 22.1 million is smaller than the first. This is restricted to polyp and breast, the skin frozen configuration not being comparable. And full fine-tuning does not lead: across the eight sets it is highest only on ETIS, so twenty-two million trainable parameters purchase no consistent advantage over fifty thousand, and this on the smallest DINOv2 variant, where the case for updating the whole encoder is strongest.

Beyond the first step the domains diverge. In polyp performance continues to rise, CVC-ColonDB gaining $+0.055$ between ranks 4 and 32; in breast and skin it does not, the corresponding changes being $-0.003$, $-0.004$ and between $+0.001$ and $+0.005$. Skin therefore requires very little adaptation of any kind, a conclusion resting on the flatness of its axis rather than on any frozen baseline. And the in-domain measurement diverges from the out-of-distribution one as it did in the budget control: in breast, rank 4 to rank 32 gains $+0.046$ on validation while both out-of-distribution sets move by less than 0.005. Additional capacity produces a model that fits the training distribution better without generalising better.

The axis reads the same in the other metrics (Table~\ref{tab:capacity}). The first step carries most of the benefit at the boundary as well: from frozen to BitFit, out-of-distribution HD95 falls by between 29 and 75 per cent on the five polyp and breast sets, from 0.196 to 0.139 on CVC-ColonDB and from 0.290 to 0.127 on ETIS, and NSD rises by between 0.13 and 0.29 on all five out-of-distribution sets. The polyp rank effect appears in every metric, CVC-ColonDB HD95 falling from 0.147 at rank 4 to 0.086 at rank 32 and NSD rising from 0.427 to 0.515; in breast, BitFit remains the best configuration out of distribution in NSD on both sets; and skin varies far less, by at most 0.05 in NSD and 0.008 in HD95 across its six configurations on any set.

Two qualifications apply to this section. BitFit received a per-domain learning-rate search whereas LoRA and full fine-tuning did not, which favours BitFit, and every row in Table~\ref{tab:capacity} is single-seed and reported as signal rather than evidence.

\FloatBarrier
\subsection{Stability of the optimal configuration}\label{sec:stability}

The preceding sections each reported a comparison; this section examines the comparisons themselves. Because the same comparisons were revisited, at two ranks, at two training budgets, against a second selection criterion, and after correcting a mismatched representational comparison, it is possible to observe which of their conclusions survive. Four comparisons were re-examined. In three an initial finding did not hold: two when a single parameter of that configuration was varied, and one when a mismatch between domains in estimator, rank and sample size was corrected. The fourth compared the same runs under a second selection criterion. A study conducted at a single operating point, which is the norm in this literature, would have reported all four in their original form.

\subsubsection{The rank artefact}

The comparison between LoRA and full fine-tuning was first conducted at rank 8, with three seeds and Wilcoxon testing, and produced a clear result: full fine-tuning led on two of three skin sets, significantly on both (20 epochs; Table~\ref{tab:rank}). At rank 32 the conclusion reverses on ISIC 2016*, from $-0.004$ ($p=0.0010$) to $+0.003$ ($p=0.0023$), and weakens on ISIC 2017* from significant to no difference, from $-0.017$ ($p=0.0004$) to $-0.002$ ($p=0.3218$), while PH2 favours LoRA at both ranks; the same reversal occurs in polyp, where CVC-ColonDB moves from $-0.024$ at rank 4 to $+0.022$ at rank 32 in the single-seed 20-epoch rank sweep. Both configurations are defensible: rank 8 is a common default and was not chosen adversarially, both comparisons used three seeds and significance testing, and both reported $p$ below 0.05 on ISIC 2016*. A reader given either in isolation, with its methodology fully described, would have no basis for doubting it. What separates them is a single hyperparameter of one of the two methods compared. The finding that full fine-tuning outperforms LoRA is not false at rank 8; it is true at rank 8 and false at rank 32, and reporting it without the rank makes it appear a statement about the methods rather than about one configuration of one of them.

\begin{table*}[tbp]
\centering
\begin{minipage}[t]{\pairwidth}
\centering
\caption{LoRA minus full fine-tuning on the three skin sets at two ranks, three seeds, 20 epochs. Positive favours LoRA.}
\label{tab:rank}
\footnotesize
\setlength{\tabcolsep}{3pt}
\begin{tabular*}{\linewidth}{@{\extracolsep{\fill}}lccccl@{}}
\toprule
Test set & $\Delta$ at $r{=}8$ & $p$ at $r{=}8$ & $\Delta$ at $r{=}32$ & $p$ at $r{=}32$ & Outcome \\
\midrule
ISIC 2016* & $-0.004$ & 0.0010 & $+0.003$ & 0.0023 & Reverses \\
ISIC 2017* & $-0.017$ & 0.0004 & $-0.002$ & 0.3218 & Significance lost \\
PH2        & $+0.0013$ & 0.0117 & $+0.003$ & $<$0.0001 & No change \\
\bottomrule
\end{tabular*}
\end{minipage}\pairsep
\begin{minipage}[t]{\pairwidth}
\centering
\caption{The four instabilities: what was varied, and the effect.}
\label{tab:instab}
\footnotesize
\begin{tabular*}{\linewidth}{@{\extracolsep{\fill}}lp{0.27\columnwidth}p{0.47\columnwidth}@{}}
\toprule
Case & What was varied & Effect \\
\midrule
Rank      & LoRA rank & Conclusion reverses on ISIC 2016* between ranks 8 and 32, and on CVC-ColonDB between ranks 4 and 32; significance lost on ISIC 2017* \\
Budget    & Epochs, 20 $\rightarrow$ 40 & OOD rank effect largely dissolves in breast, persists in polyp \\
Selection & Nothing; post-hoc & Validation-optimal $\neq$ OOD-optimal in five of six searches \\
\begin{tabular}[t]{@{}l@{}}Mismatched\\comparison\end{tabular} & CKA estimator, rank and sample size & Direction of representational drift reverses \\
\bottomrule
\end{tabular*}
\end{minipage}
\end{table*}

\subsubsection{The selection-criterion artefact}

Selection was performed on in-domain validation DSC throughout, with out-of-distribution scores recorded but unused, which makes it possible to ask after the fact whether the configuration selected was the one best out of distribution. In six independent searches, only one did. In the BitFit learning-rate search in polyp at 40 epochs (Table~\ref{tab:lrsearch}), validation selects 1e-3, best on three of five test sets but not on CVC-ClinicDB, where 1e-2 leads by 0.005, or CVC-ColonDB, where 3e-3 leads by 0.009. The plain U-Net search behaves identically, validation selecting 1e-3 while CVC-ColonDB and CVC-T are better served by 3e-3. The BitFit search in skin diverges on one of its three test sets, PH2 being better served by 3e-4 by 0.008. Two remaining instances are of the same kind: in the ResU-Net search validation selected 3e-3 while ETIS was best served by 1e-2, and in the breast rank sweep in-domain performance rose monotonically with rank while out-of-distribution performance did not. The sixth search, BitFit in breast, is the exception: validation selected 3e-3, which was also best on both out-of-distribution sets, at 0.7337 on BrEaST and 0.7449 on BUS-UCLM.

\begin{table*}[tbp]
\centering
\caption{BitFit learning-rate search, polyp, 40 epochs. Validation selects 1e-3. Bold marks the best value in each column.}
\label{tab:lrsearch}
\footnotesize
\begin{tabular*}{\textwidth}{@{\extracolsep{\fill}}lcccccc@{}}
\toprule
Learning rate & Val & Kvasir & ClinicDB & ColonDB & ETIS & CVC-T \\
\midrule
1e-2 & 0.8866 & 0.8480 & \textbf{0.8359} & 0.7256 & 0.6509 & 0.8558 \\
3e-3 & 0.8712 & 0.8503 & 0.7691 & \textbf{0.7293} & 0.5689 & 0.8273 \\
1e-3 & \textbf{0.8958} & \textbf{0.8960} & 0.8306 & 0.7204 & \textbf{0.6867} & \textbf{0.8634} \\
3e-4 & 0.8826 & 0.8731 & 0.8158 & 0.6892 & 0.6604 & 0.8587 \\
\bottomrule
\end{tabular*}
\end{table*}

The criterion applied here is rank order on individual test sets: in five of six independent searches, across two hyperparameters and two model families, the validation-optimal configuration was not the out-of-distribution-optimal one on at least one test set. Under a stricter criterion the picture is weaker. Averaged over the three polyp out-of-distribution sets the validation choice is also the best of the four rates, at 0.757 against 0.744 for the nearest alternative, and the two individual divergences of 0.005 and 0.009 fall below the floor; in skin the same criterion costs 0.002, again within it. What the searches establish is therefore that the validation optimum does not coincide with the optimum on every deployment target, not that selecting on validation carries a measurable performance penalty. The direction of the misalignment is structural, since selection must use data available at training time, which is in-domain by construction; its magnitude is small and was not the object of this study.

\subsubsection{The mismatched-comparison artefact}

Centred kernel alignment \citep{kornblith2019cka} was used to measure how far each strategy moves the encoder's representations from their pretrained state, and the initial analysis produced an interpretable result: LoRA preserved the early blocks while full fine-tuning perturbed them throughout, fitting the expectation that a constrained update should stay closer to its initialisation. But three variables differed between the domains analysed, polyp using the biased estimator at rank 32 on one sample size and skin the debiased estimator at rank 8 on another. Repeated under one consistent procedure, the debiased estimator of \citet{nguyen2021wide} at matched ranks with fixed sample size, the result reverses: LoRA drifts further from the pretrained representation than full fine-tuning, in both domains and at both ranks, with no exception at any block. The estimator was not the cause; the reversal came from the mismatched rank and sample size. The corrected direction is explained by layer-wise decay, which holds full fine-tuning's lower layers close to initialisation by construction while LoRA applies a uniform rate at every depth. Unlike the first two cases, the original analysis was simply wrong rather than conditional on a defensible choice; it is reported because the corrected finding is used in Section~\ref{sec:predictors}, and because the erroneous version was interpretable and fitted theoretical expectation.

\subsubsection{Synthesis}

The fourth instability, the epoch-budget artefact, appeared in Section~\ref{sec:capacity} because it determined the budget used thereafter; Table~\ref{tab:instab} summarises all four. The four cases are independent, arising in different experiments. Two follow from varying an operating point, the rank and the budget; the other two are a post-hoc comparison of two selection criteria on the same runs and the correction of a comparison whose estimator, rank and sample size had differed between domains. What they share is that each was interpretable and internally consistent before it was examined a second time: one was supported by significance testing across three seeds, the two operating-point findings were undetectable from within the original experiment, and the mismatched comparison was detectable only by checking its procedure.

\FloatBarrier
\subsection{Candidate explanations for the domain effect}\label{sec:predictors}

The deficit of task-specific training varies by domain and not by architecture or training-set size, and the capacity required varies by domain and not by method. In both cases skin is the least demanding, though the two axes differ on the ordering of polyp and breast. If it could be predicted from a measurable property, a practitioner could estimate in advance whether a foundation model is warranted and what budget it requires. Three analyses were run. Two test candidate predictors of the domain effect: distributional distance between training and test data, and the spectral dimension of the fine-tuning update. The third asks a related question, whether the difference between the two adaptation methods lies in the dimensionality of the solutions they reach. None succeeded, and the negative results are reported because the failure of the obvious explanations is itself informative.

Maximum mean discrepancy \citep{gretton2012kernel} was computed between the training set and each test set on features from the pretrained encoder. The measure behaves sensibly on its own terms, the in-domain sets being closest and the estimator returning 0.0380 against a resampled training split and 0.0001 against a shuffled control. The prediction nonetheless fails, and on the most extreme point: CVC-T has the highest MMD$^2$ at 0.200, nearly twice CVC-ColonDB's 0.113, yet is the easiest out-of-distribution set at 0.867 under LoRA $r{=}32$ (20 epochs, three-seed mean), twelve points above CVC-ColonDB and seventeen above ETIS. Since every CVC-T image also occurs in CVC-ColonDB (Section~\ref{sec:datasets}), this compares a subset of one collection with the whole of it, which sharpens the dissociation: the subset lies further from the training data than the collection it is drawn from, and is easier. Skin gives the same picture in miniature: PH2 has the largest MMD$^2$ at 0.0739, an order of magnitude above ISIC 2017* at 0.0070, and is not correspondingly harder. Distance in the pretrained feature space is not difficulty, since a set can be far from the training distribution in a way the encoder registers while remaining easy to segment.

The second candidate addresses the adaptation axis. If a domain requires a higher rank, the update full fine-tuning finds might itself be higher-dimensional, and the singular value spectrum of that update was computed across all encoder blocks. Measured as the number of singular directions capturing 90 per cent of the update's energy, it is 28.4 in skin, 39.0 in breast and 42.1 in polyp at 20 epochs (Table~\ref{tab:spectral}). At 20 epochs the measure tracks the ordering, skin lowest and polyp highest, but two problems prevent it being a usable predictor. It separates the domains far less than the rank sweep does, spectral dimension spanning a factor of 1.5 against eight for rank, and breast sits closer to polyp than to skin on this measure while behaving like skin on the capacity axis at the extended budget. And the measure is budget-sensitive, breast rising from 39.0 to 44.3 between 20 and 40 epochs, a change larger than the gap between breast and polyp at either budget, so it is not intrinsic to the domain: the epoch-budget artefact in a second guise.

\begin{table}[tbp]
\centering
\caption{Spectral properties of the full fine-tuning update. The reported quantity is the number of singular directions capturing 90 per cent of the update's energy. Polyp and skin at 20 epochs.}
\label{tab:spectral}
\footnotesize
\setlength{\tabcolsep}{3pt}
\begin{tabular*}{\columnwidth}{@{\extracolsep{\fill}}lcccc@{}}
\toprule
Domain & Directions at 90\% & at 95\% & at 99\% & Effective rank \\
\midrule
Skin               & 28.4 & 48.5 & 113.2 & 21.3 \\
\midrule
Breast (20 epochs) & 39.0 & 65.1 & 139.3 & 26.1 \\
\midrule
Polyp              & 42.1 & 69.8 & 146.2 & 27.0 \\
\midrule
Breast (40 epochs) & 44.3 & 71.7 & 145.5 & 30.7 \\
\bottomrule
\end{tabular*}
\end{table}

The third analysis asks whether the difference between LoRA and full fine-tuning lies in the dimensionality of the solution each finds. The full fine-tuning update was truncated by singular value decomposition across all modified two-dimensional encoder weight matrices, with one-dimensional parameters kept at their fine-tuned values, and added back to the pretrained weights (polyp, 20 epochs, seed 42; Table~\ref{tab:trunc}). Truncating to rank 32 costs at most 0.007 on any set, and at rank 4 the two methods separate slightly, LoRA leading on CVC-ClinicDB by $+0.033$. The full fine-tuning update examined can therefore be approximated at low rank, though the truncation covers every modified encoder weight matrix while LoRA modifies only the query and value projections, so the two solution spaces are not identical.

\begin{table*}[tbp]
\centering
\caption{Truncated full fine-tuning against LoRA, polyp, 20 epochs, seed 42 (single seed). Truncation columns report the update projected onto its leading $r$ singular directions; LoRA columns an adapter trained at the same rank. These are 20-epoch checkpoints, distinct from the 40-epoch runs of Table~\ref{tab:capacity}; the LoRA $r{=}32$ run is the seed-42 member of the three-seed mean in Table~\ref{tab:lorafull}.}
\label{tab:trunc}
\footnotesize
\begin{tabular*}{\textwidth}{@{\extracolsep{\fill}}lccccc@{}}
\toprule
Test set & Full FT & Trunc $r{=}4$ & LoRA $r{=}4$ & Trunc $r{=}32$ & LoRA $r{=}32$ \\
\midrule
Kvasir       & 0.8753 & 0.8638 & 0.8609 & 0.8743 & 0.8765 \\
CVC-ClinicDB & 0.8369 & 0.7838 & 0.8166 & 0.8295 & 0.8481 \\
CVC-ColonDB  & 0.7482 & 0.7244 & 0.7237 & 0.7469 & 0.7700 \\
ETIS         & 0.6983 & 0.6610 & 0.6582 & 0.6957 & 0.6961 \\
CVC-T        & 0.8690 & 0.8561 & 0.8701 & 0.8682 & 0.8578 \\
\bottomrule
\end{tabular*}
\end{table*}

Two measurements were taken as candidate predictors and neither yields a usable rule, one failing to track the ordering and the other tracking it at one budget but moving with the budget; the truncation analysis addresses a different question, and the corrected analysis of Section~\ref{sec:stability} adds a further negative result, ruling out the account under which LoRA generalises better by staying closer to its initialisation. The domain-dependence is therefore an empirical regularity established across two axes and, on the capacity axis, two methods sharing no mechanism, and it remains without a mechanistic account.

\section{Discussion}\label{sec:discussion}

\subsection{Adaptation}\label{sec:insights}

Four findings concern adaptation itself: where it helps, what it costs, how much of it is needed, and how far comparisons of it can be trusted.
\begin{enumerate}
\item \textit{The advantage of adaptation lies out of distribution.} Where the deployment target is the training distribution and the full training set is available, no accuracy advantage for adaptation was observed. Averaged over the two in-domain polyp test sets the plain U-Net scores within the measurement floor of the adapted model, at fewer parameters and no pretraining, and in breast both task-specific models score above it on the validation split, though that split also served for model selection. On the smaller polyp subsets adaptation does lead. Where the target is data from a different institution, scanner or protocol, adaptation leads on every set examined: all sixteen comparisons across eight sets and two independently tuned architectures, although the two on CVC-T repeat those on CVC-ColonDB rather than adding independent evidence. Recomputed without CVC-T, the polyp out-of-distribution means fall, but no ordering, sign or trend reported in Section~\ref{sec:results} changes (Table~\ref{tab:cvct}). The margin varies by more than a factor of ten and is conditional on training-set size in a specific way: within a domain more data reduces it, but across domains data volume does not predict it.
\item \textit{The advantage has a cost in small lesions and fine boundaries.} It is not uniform across metrics and kinds of error (Section~\ref{sec:failure}): it holds in overlap, sensitivity, precision and HD95, but at the finest boundary scale the task-specific models are as good or better at 224 pixels, and the adapted model more often misses a small lesion altogether, 13 polyp images at the selected configuration against none for the plain U-Net.
\item \textit{Little adaptation is needed.} The step from a frozen encoder to BitFit captures most of the available benefit in polyp and breast, and beyond it LoRA at 2.7 per cent of the parameters is equal or ahead of full fine-tuning in DSC, IoU and NSD on all five polyp sets across three seeds. The equality does not extend to every metric: LoRA's precision is significantly lower on all three polyp out-of-distribution sets, so at equal DSC it covers more of each lesion and more of its surroundings.
\item \textit{In-domain orderings do not reliably carry over, and conclusions depend on the operating point.} Between architectures (Section~\ref{sec:main}), between data scales (Section~\ref{sec:datasize}) and along the adaptation axis (Section~\ref{sec:capacity}), the ordering or optimum identified on held-out data from the training distribution did not reliably carry over to data from a different source. The budget control of Section~\ref{sec:budget} separates the two regimes within a single experiment: doubling the epoch budget in breast largely dissolved the out-of-distribution rank effect while leaving the in-domain effect intact. Of the four cases in Section~\ref{sec:stability}, three revised a finding: two reversed or weakened when the LoRA rank or the training budget was changed, one of them despite significance testing across three seeds, and one reversed once a mismatched comparison was corrected; in each the original version claimed the larger effect. The fourth compared two selection criteria post hoc on the same runs.
\end{enumerate}

\begin{table*}[tbp]
\centering
\caption{Polyp out-of-distribution means with and without CVC-T, whose 60 images are pixel-identical to images in CVC-ColonDB (Section~\ref{sec:datasets}). Means are unweighted averages over test sets: CVC-ColonDB, ETIS and CVC-T (with) or CVC-ColonDB and ETIS (without). Deficits and gaps are task-specific model minus DINOv2 + LoRA $r{=}32$. Recomputed from the same checkpoints as the main text; BitFit learning rates from the per-set values of Table~\ref{tab:lrsearch}.}
\label{tab:cvct}
\footnotesize
\begin{tabular*}{\textwidth}{@{\extracolsep{\fill}}llcc@{}}
\toprule
Quantity & Configuration & With CVC-T & Without CVC-T \\
\midrule
Mean DSC (Table~\ref{tab:main})          & Plain U-Net        & 0.7370 & 0.6852 \\
                            & ResU-Net           & 0.7120 & 0.6709 \\
                            & LoRA $r{=}32$      & 0.7781 & 0.7339 \\
\midrule
Deficit (Table~\ref{tab:deficit})          & Plain U-Net        & $-0.041$ & $-0.049$ \\
                            & ResU-Net           & $-0.066$ & $-0.063$ \\
\midrule
OOD gap, data-size sweep (Table~\ref{tab:datasize}) & 10\%        & $-0.191$ & $-0.178$ \\
                            & 25\%               & $-0.135$ & $-0.154$ \\
                            & 50\%               & $-0.132$ & $-0.135$ \\
                            & 100\%              & $-0.066$ & $-0.063$ \\
\midrule
Mean DSC, capacity axis (Fig.~\ref{fig:capacity}) & Frozen        & 0.6143 & 0.5411 \\
                            & BitFit             & 0.7568 & 0.7035 \\
                            & LoRA $r{=}4$       & 0.7552 & 0.6991 \\
                            & LoRA $r{=}8$       & 0.7545 & 0.7058 \\
                            & LoRA $r{=}16$      & 0.7602 & 0.7028 \\
                            & LoRA $r{=}32$      & 0.7781 & 0.7339 \\
                            & Full FT            & 0.7725 & 0.7218 \\
\midrule
Mean DSC, BitFit search (Table~\ref{tab:lrsearch}) & 1e-2       & 0.7441 & 0.6883 \\
                            & 3e-3               & 0.7085 & 0.6491 \\
                            & 1e-3 (selected)    & 0.7568 & 0.7036 \\
                            & 3e-4               & 0.7361 & 0.6748 \\
\bottomrule
\end{tabular*}
\end{table*}

\subsection{Domain}\label{sec:interpret}

The clearest result of this work is that the amount of adaptation required belongs to the domain rather than to the method. The rank at which adaptation ceases to improve varies across the three domains, polyp continuing to improve to rank 32, breast plateauing at 16 and skin showing no movement beyond rank 4 under the 20-epoch protocol, though Section~\ref{sec:budget} shows the breast effect largely dissolves out of distribution at twice that budget. Polyp is also the one domain where BitFit falls behind, while skin and breast are served by the smallest budget tested under both methods. That two methods differing by a factor of eleven in parameter count, in intervention site, and in whether they possess a rank parameter at all place the same domain above the others locates the requirement in the domain. Other configuration choices behave the same way: the learning rate selected on validation differed between architectures and between domains, and the optimal budget differed by domain, so a study on any one of these domains would have produced an internally consistent set of settings that does not transfer to the other two.

The architecture axis places the domains in a related but not identical order. Both axes separate skin from the other two, skin showing the smallest deficit of task-specific training as well as the smallest capacity requirement, but they differ on polyp and breast, the mean deficit in Table~\ref{tab:deficit} being larger in breast than in polyp under both architectures. Four explanations for the domain-dependence can be set aside. It is not the task-specific architecture, since the deficit holds under two such models with identical sign on all eight sets, the simpler being the stronger out of distribution. It is not the method, since the ordering holds under LoRA and BitFit, which share no mechanism. It is not distance in the pretrained feature space, since the polyp set with the highest discrepancy is the easiest. And it is not the encoder family, since the same transformer without adaptation reaches 0.614 out of distribution in polyp against 0.737 for the plain U-Net and in breast falls within the measurement floor of the plain U-Net, so the advantage appears only once adaptation is applied. Architecture and pretraining remain confounded, since no convolutional encoder pretrained at comparable scale was included, but the frozen result shows the encoder family alone does not produce the effect. Training-set size is a fifth candidate and is only partly addressed: on the architecture axis the deficit does not follow a simple ordering by size, skin at 518 images showing a much smaller gap than polyp at 652.

The ordering also conceals a difference in what kind of shift each test set represents. Two of three skin sets are earlier releases of the training archive, suggesting skin's small deficit reflects a milder shift rather than a property of dermoscopy. The prediction that follows fails: PH2, the only independent skin source, should show the largest skin deficit and under the plain U-Net shows the smallest. Within-domain variation is of comparable magnitude, so treating the domain as the unit of analysis is a simplification the test-set-level results only partly support.

The account most consistent with the results is that domains differ in how much of the required update is task-general rather than domain-specific, and that pretraining supplies the former. A foundation model arrives with representations encoding much of what segmentation requires, and adaptation supplies only the domain-specific remainder. Where that remainder is small, little capacity is needed and pretraining transfers well; where it is large, more is needed, and a model trained from random initialisation must learn the general component from target data alone, which it can do well enough to fit the training distribution but not well enough to transfer. This fits four observations: why the first step on the capacity axis carries most of the benefit, why the requirement is method-independent, why task-specific models are competitive in domain but not out of it, and why more data narrows the in-domain gap faster. It was not tested, and Section~\ref{sec:predictors} records the closest attempt.

That the plain variant outperforms the residual one is itself unexpected, and the most plausible reconciliation is scale. Residual connections address optimisation difficulty in deep networks, and at five encoder levels and under eight million parameters that difficulty may not arise, leaving the additional skip projections as parameters that fit the training distribution without contributing out of distribution. This is untested here.

\subsection{Practical recommendations}\label{sec:recommendations}

The first four recommendations are for practitioners choosing between the approaches, the last two for studies comparing them.
\begin{enumerate}
\item \textit{Use an adapted foundation model when the model will be deployed on data from other sites, scanners or protocols.} It led on all eight out-of-distribution test sets under both task-specific architectures, in DSC, IoU, sensitivity and HD95, and its lead grows as training data shrink, most clearly in polyp. When deployment stays within the training source and the full training set is available, a well-configured U-Net is sufficient: no accuracy advantage for adaptation was observed there, and the case for a foundation model must then rest on engineering effort or on one backbone serving several tasks.
\item \textit{Start with the smallest adaptation and do not default to full fine-tuning.} Moving from a frozen encoder to BitFit captures most of the benefit in polyp and breast, and LoRA at 2.7 per cent of the parameters equalled or exceeded full fine-tuning in DSC on all five polyp sets.
\item \textit{Sweep the adaptation capacity on the target domain instead of inheriting it.} The LoRA rank required ranged from 4 in skin to 32 in polyp, and the same domain pattern appeared under BitFit, so a value tuned on another dataset is not a valid default.
\item \textit{Check small lesions and boundaries explicitly before deployment.} At 224 pixels the adapted model missed small polyps entirely where the U-Nets did not, and a U-Net traced boundaries more precisely on four of the five polyp and breast out-of-distribution sets. Where small lesions or exact boundaries matter, report sensitivity, empty predictions and NSD alongside DSC, and evaluate at the highest resolution the budget allows: at 392 pixels the adapted model led in NSD on all three skin sets.
\item \textit{Report in-domain and out-of-distribution results separately.} An in-domain result is not evidence for cross-site deployment: on all three axes studied here the in-domain ordering did not reliably carry over out of distribution, and in Section~\ref{sec:budget} the two regimes diverged within a single experiment.
\item \textit{Establish a measurement floor, and vary the operating point, before reporting a difference as a property of a method.} Repeated runs at an identical seed differed by up to 0.013 DSC, of the same order as the improvements this literature commonly reports, and two findings here reversed or weakened when the LoRA rank or the training budget was changed; where a comparison involves a hyperparameter of one method, establish the conclusion at more than one value of it.
\end{enumerate}

\subsection{Limitations}\label{sec:limitations}

Seven limitations bound these recommendations, each with the study that would remove it.
\begin{enumerate}
\item \textit{A single backbone.} All four adaptation strategies were applied to DINOv2-Small, so while the capacity requirement is shown not to be specific to the adaptation method, whether it is specific to the backbone was not tested, and these results do not recommend DINOv2 over other backbones. Repeating the adaptation axis on a second encoder such as SAM under an identical protocol would settle this.
\item \textit{Asymmetric tuning.} BitFit received a per-domain learning-rate search while LoRA and full fine-tuning used rates held constant from earlier experiments, which favours BitFit, and part of its standing may reflect that search. The task-specific learning-rate search was conducted on polyp alone, and in skin the plain variant ran at the residual variant's rate.
\item \textit{Mostly single-seed results.} Only the core comparison between LoRA and full fine-tuning used three seeds and significance testing; the capacity axis, BitFit, the plain U-Net and the data-size sweep are reported as signal and read against the measurement floor of Section~\ref{sec:eval}. Three configurations run three times each would give a per-domain variance estimate.
\item \textit{Three domains.} They establish that the capacity requirement varies by domain, but not what determines it or how to anticipate it for a fourth domain without its data; answering that requires more domains under one protocol. Whether the requirement also varies with training-set size was not tested, since the rank sweep used one training-set size per domain, and the domain ordering should not be extrapolated beyond the three domains studied.
\item \textit{Training-set range.} In-domain parity was observed at the full training sets, of 479 to 2,075 images, and in breast down to 119 images, while on the smaller polyp subsets adaptation led; where parity breaks down was not located, so these results do not show pretraining unnecessary in domain in general.
\item \textit{A minimal decoder.} The linear decoder isolates encoder adaptation but makes the frozen configuration a minimal baseline; repeating the frozen and BitFit configurations with a convolutional decoder would separate the decoder's contribution.
\item \textit{The breast in-domain figure.} It comes from the validation split, which was also used for early stopping, and is therefore optimistically biased. Polyp's in-domain sets are separate from the validation split, and skin has no in-domain figure.
\end{enumerate}

\subsection{Relation to previous work}\label{sec:relation}

\textit{Evaluation hazards in standard benchmarks.} Two hazards in the data used here were found by content-based checks that file-hash comparison does not replicate. In the polyp protocol, a pixel-level comparison shows that all 60 CVC-T images are identical to images in CVC-ColonDB, as are 59 of their 60 masks, while no file is byte-identical, so an md5 check passes; this is consistent with EndoScene defining its images as drawn from CVC-ColonDB \citep{vazquez2017benchmark}. Studies following the standard protocol therefore count these images twice when reporting generalisation. In the breast training source, a perceptual hash at a Hamming distance of five or less placed 212 of the 647 lesion-bearing BUSI images in 98 near-duplicate groups, only one of which contains byte-identical files; keeping one image per group removed 114 images (17.6 per cent of the 647), and inspection of a sample of four groups showed the same acquisitions annotated twice with differing masks. The count depends on the threshold, from 89 images (13.8 per cent) at a distance of three to 137 (21.2 per cent) at eight. All 114 were removed before any experiment reported here.

That LoRA matches full fine-tuning at a fraction of the parameters is established outside medical imaging and is confirmed here, and the BitFit result extends a claim reported for language tasks in low-data regimes to dense prediction on medical images and to cross-dataset evaluation, neither of which the original work or its medical successors examine.

More interesting is the disagreement of Section~\ref{sec:rw-peft}. \citet{dutt2024peft} attribute the ordering among methods to data volume; \citet{teuber2026peft} attribute it to dataset characteristics. These results support the second, and support the explanation it offers for the disagreement: Section~\ref{sec:capacity} finds BitFit ahead of LoRA on both breast out-of-distribution sets, level on all three skin sets, and behind on one polyp set, a ranking that changes with the dataset rather than holding globally. A study on any one of these domains would report a different ordering, each correct about its own, so the disagreement need not be an error on either side. One difference in scope should be noted alongside the agreement. Their base model is SAM, supervised on segmentation masks, whereas DINOv2 has never seen one, so the amount of adaptation each requires is not the same quantity; and their evaluation does not separate in-domain from cross-dataset performance, whereas that divergence is the central finding here.

\citet{isensee2024revisited} report strong performance from residual-encoder configurations, which motivated the residual variant used here. Under the conditions of this study, that variant, with residual blocks in both encoder and decoder, did not outperform its plain counterpart. Because it is not the nnU-Net ResEnc configuration and the setting differs, three-dimensional and multi-class on larger datasets there, two-dimensional and binary here, the result does not test their recommendation directly.

The literature on foundation models for medical segmentation generally reports adapted models outperforming task-specific training. These results agree out of distribution and disagree in domain, finding no advantage in either domain where an in-domain comparison is possible, with parity reached at 119 training images in breast, against baselines Section~\ref{sec:arch} shows competitive with published architectures. The discrepancy is most likely one of evaluation regime rather than of result: studies reporting an advantage frequently evaluate on data from the training distribution or do not distinguish the regimes, and where the distinction is drawn the advantage concentrates entirely out of distribution. This locates the existing result more precisely rather than contradicting it.

Independent support comes from breadth rather than depth. U-Bench \citep{tang2025ubench} reports that improvements under zero-shot generalisation are more consistently significant than those measured in domain, which is the same divergence between regimes observed from the opposite direction: where this work finds an in-domain measurement failing to predict out-of-distribution behaviour, U-Bench finds the out-of-distribution measurement separating methods the in-domain one does not.

Finally, the comparison drawn throughout this study is binary, and a recent line of work pursues a third option: distilling foundation-model representations into a lightweight architecture so that the result carries those priors at the parameter count of a U-Net. Polyp-DiFoM \citep{agnihotri2026difom} distils SAM, DINOv2, OneFormer and Mask2Former into U-Net students and evaluates on the same five polyp sets used here. This follows from the present comparison rather than competing with it: if the advantage is concentrated out of distribution and a lightweight architecture already matches in domain, transferring that component into the smaller model is the natural next step, and the domain dependence found here implies that a distillation method validated on one domain should not be assumed to transfer either.

\section{Conclusion}\label{sec:conclusion}


This study compared foundation-model adaptation with task-specific training from scratch for medical image segmentation, separating in-domain from out-of-distribution evaluation across three domains and eight out-of-distribution test sets. Three conclusions follow.
\begin{enumerate}
\item Adaptation is warranted when the deployment target differs from the training source: an adapted DINOv2 outperformed two independently tuned U-Nets on all eight out-of-distribution sets in DSC, IoU, sensitivity and HD95, whereas within the training distribution and at the full training set a well-configured U-Net matched it.
\item Little adaptation is needed: bias-only tuning captures most of the benefit, and LoRA at 2.7 per cent of the parameters equals or exceeds full fine-tuning in DSC on all five polyp sets.
\item How much adaptation is needed is a property of the domain rather than of the method, varying eightfold in LoRA rank across domains and reproduced by a method with no rank parameter, so capacity must be selected on the target domain.
\end{enumerate}
These results were obtained with DINOv2, a small, general-purpose foundation model never trained on segmentation masks or medical images; larger or medical-specific foundation models would be expected to perform at least as well.

\section*{Data availability}
All datasets used in this study are publicly available and were released for research use by their original providers: Kvasir-SEG \citep{jha2020kvasir}, CVC-ClinicDB \citep{bernal2015wm}, CVC-ColonDB \citep{bernal2012polyp}, ETIS-LaribPolypDB \citep{silva2014etis} and CVC-T \citep{vazquez2017benchmark}, used under the protocol of \citet{fan2020pranet}; BUSI \citep{aldhabyani2020busi}, BrEaST \citep{pawlowska2024breast} and BUS-UCLM \citep{vallez2025busuclm}; and ISIC 2016 \citep{gutman2016isic}, ISIC 2017 \citep{codella2018isic}, ISIC 2018 \citep{codella2019isic, tschandl2018ham} and PH2 \citep{mendonca2013ph2}.

\section*{Code availability}
The code used for all experiments and analyses, together with the per-image metrics for every evaluated model, is available at \url{https://github.com/emrekutlusan/foundation-model-vs-scratch-medical-segmentation} under the MIT licence.

\section*{Ethics statement}
All datasets used here are publicly available and were released for research use by their original providers, and no new human data were collected.

\section*{Acknowledgements}
The authors thank the authors and clinical teams behind the public datasets used here, whose work made a study of this kind possible.

\section*{Funding}
This research did not receive any specific grant from funding agencies in the public, commercial, or not-for-profit sectors.

\section*{CRediT authorship contribution statement}
\textbf{EK:} Conceptualization, Methodology, Software, Validation, Formal analysis, Investigation, Data curation, Visualization, Writing -- original draft, Writing -- review \& editing.
\textbf{YL:} Conceptualization, Methodology, Supervision, Project administration, Resources, Writing -- review \& editing.

\section*{Declaration of competing interest}
The authors declare that they have no known competing financial interests or personal relationships that could have appeared to influence the work reported in this paper.

\section*{Declaration of generative AI and AI-assisted technologies in the manuscript preparation process}
Generative AI tools, including Claude (Anthropic), were used in a supporting role. During the underlying project they were used for checking grammar and proof-reading, for drafting and revising passages of the text from content the authors provided, and for assistance in debugging analysis code and producing figures from numerical results that the authors supplied and verified. In preparing this manuscript, they were used to help restructure and condense the text, to assist in writing the extended evaluation and analysis code to the authors' specification, and to check reported numbers and cited statements against their sources. The study design, data preparation, model training, experiments, verification of all results and their interpretation are the authors' own; the authors reviewed and edited all content and code and take full responsibility for them.

\bibliographystyle{elsarticle-num-names}  
\bibliography{refs}

\end{document}